\documentclass[preprint,3p]{elsarticle}

\usepackage{graphicx}
\usepackage{amsmath,amssymb,amsfonts}
\usepackage{booktabs}
\usepackage{array}
\usepackage{tabularx}
\usepackage{xcolor}
\usepackage{url}
\usepackage[hidelinks]{hyperref}
\usepackage{rotating}
\usepackage{xspace}

\usepackage{fontspec}
\newfontfamily\bnfont{kalpurush.ttf}[Script=Bengali]
\newcommand{\bn}[1]{{\bnfont #1}}

\journal{Computer Vision and Image Understanding}
\graphicspath{{Figures/}}

\newcommand{\nameModel}{BarkNet-Lite\xspace}
\newcommand{\nameDataset}{BarkBD\xspace}

\newcommand{\tim}{\textsc{tim}\xspace}   
\newcommand{\tam}{\textsc{tam}\xspace}   
\newcommand{\trb}{\textsc{trb}\xspace}   
\newcommand{\cam}{\textsc{cam}\xspace}   

\newcommand{\R}{\mathbb{R}}
\DeclareMathOperator*{\argmax}{arg\,max}
\DeclareMathOperator{\GAP}{GAP}
\DeclareMathOperator{\GMP}{GMP}
\DeclareMathOperator{\BN}{BN}
\DeclareMathOperator{\Concat}{Concat}

\begin{document}

\begin{frontmatter}
\title{\nameModel: A Lightweight Texture and Colour Network with the \nameDataset Benchmark for Bark-Based Tree Species Recognition in Bangladesh}

\author[kuet]{Aroshi Ali\corref{cor1}}
\ead{aroshiali28@gmail.com}

\author[baust]{Saad Ahmed}
\cortext[cor1]{Corresponding author.}

\author[baust]{Md. Khalid Syfullah}

\affiliation[kuet]{
    organization={Department of Computer Science and Engineering, Khulna University of Engineering \& Technology},
    city={Khulna},
    postcode={9203},
    country={Bangladesh}
}

\affiliation[baust]{
    organization={Department of Computer Science and Engineering, Bangladesh Army University of Science \& Technology},
    city={Saidpur},
    country={Bangladesh}
}

\begin{abstract}
Tree species recognition supports forest inventory and biodiversity monitoring but still depends on scarce taxonomic expertise. Bark is visible year-round at ground level, yet bark recognition has concentrated on temperate floras and on large ImageNet-pre-trained backbones. We address both gaps. First, we release \nameDataset, a bark dataset for Bangladesh: 14{,}258 uncropped smartphone photographs of 20 native species across four districts and three weather conditions, with a fixed stratified split. Second, we propose \nameModel, a 2.96\,M-parameter network trained from random initialisation, pairing a multi-scale texture pathway with a parallel colour-aware pathway. Over five seeds it reaches $96.64 \pm 0.66\%$ accuracy under strict single-image inference, within 2.3 points of nine ImageNet-pre-trained backbones fine-tuned under an identical protocol and within one seed-level standard deviation of the smallest of them, and transfers to public benchmarks (95.86\% on BarkVN-50, 92.85\% on BarkNet 1.0). Grad-CAM, validated by faithfulness and weight-randomisation checks, confirms its decisions rest on bark structure rather than background. The exported single-precision model classifies one photograph in $15.34$\,ms on a commodity smartphone.
\end{abstract}


\end{frontmatter}

\section{Introduction}
\label{sec:introduction}

Managing forests begins with species identity: global assessments, national inventories, urban tree registries, timber-legality audits and carbon-credit verification all rest on species-level records, each bottlenecked by one scarce resource---a trained observer who can name a tree in the field \cite{fao2020global}. That observer is scarcest in the tropics, where species richness is highest and taxonomic capacity lowest. Bangladesh is a clear case: most of its trees stand outside gazetted forest land, on roadsides and in homesteads and urban parks, and inventory must be repeated often as land use changes quickly \cite{bfd2019bangladesh}, making manual identification at that cadence infeasible.

Automating identification from images is therefore attractive, and plant recognition is now one of applied computer vision's visible successes, with citizen-science platforms handling millions of requests a year \cite{joly2016look}. Almost all of it rests on leaves, flowers and fruit, which are information-rich but seasonally unavailable and often out of reach in a mature canopy. Bark is neither: present year-round and at eye level regardless of tree height, it is the natural substrate for a tool that must work wherever a surveyor stands.

Bark is also a hard signal. Its discriminative content lies in fine-grained texture---fissure depth and orientation, lenticel arrangement, exfoliation geometry---rather than in the global shape that dominates object recognition. Intra-class variation is large, since bark changes with age and diameter, while inter-class variation can be small, since unrelated species converge on similar textures; and the same trunk in sun, cloud and rain looks markedly different, moisture darkening and saturating colour in ways no colour-constancy heuristic fully undoes.

Research has responded in two waves. The first used handcrafted texture descriptors, interpretable and cheap but plateauing below operational accuracy \cite{fekri2020bark,boudra2017statistical}. The second learned the representation: convolutional networks on purpose-built corpora now exceed 90\% on temperate species \cite{carpentier2018tree,wu2021deep,kim2022identifying}, and the field has consolidated around a handful of public benchmarks \cite{carpentier2018tree,warner2024centralbark,faizal2022automated,ratajczak2019efficient,svab2014trunk12,lakmann1998barktex}.

Two gaps motivate this work. The first is geographic. Established benchmarks come from North America, Europe and East Asia; the only South Asian corpus we are aware of, BarkVisionAI \cite{chhatre2026barkvisionai}, covers 13 species from two Indian states under a reuse-restricting licence. The Bengal delta is absent entirely: mango, jackfruit, mahogany, betel nut and the roadside palms appear in no public bark dataset, so the target classes exist in no source label space and only new collection can supply them.

The second gap is architectural. The accuracies above are almost always obtained by fine-tuning an ImageNet-pre-trained backbone with tens of millions of parameters, which assumes that a suitable licensable checkpoint exists, that its object-centric inductive biases suit a texture problem, and that the result fits on the device a surveyor carries. Knowledge distillation \cite{hinton2015distilling} shrinks such models after the fact \cite{wu2021deep}, accepting the deployment constraint without questioning the design that created it. A compact architecture designed around the signal and trained from scratch has not been systematically evaluated for bark, and the question is not whether it beats a large pre-trained model but how much accuracy independence costs.

We pursue both directions. We construct \nameDataset, a bark dataset for Bangladesh: 14{,}258 uncropped smartphone photographs of 20 native species across four districts spanning the northern plains and south-eastern coast, under sunny, cloudy and rainy conditions, kept exactly as captured. On it we propose \nameModel, built on one hypothesis: bark identity is carried by texture at several scales at once and, more weakly, by colour (the grey-green of \textit{Eucalyptus}, the red-brown of \textit{Swietenia}), which standard deep stacks entangle and discard through downsampling. \nameModel keeps the two apart: a texture pathway of inception-style multi-scale blocks, channel-and-spatial attention and residual bottlenecks, and a parallel colour-aware pathway tapped early and reinjected at two later depths. The network holds 2.96 million parameters and is trained from random initialisation.

\subsection{Research questions}
\label{subsec:rqs}

\begin{itemize}
\item\textbf{RQ1:} Where does a purpose-built lightweight network trained from scratch sit on the accuracy--cost frontier, relative to modern ImageNet-pre-trained backbones fine-tuned on the same regional imagery under an identical protocol?
\item\textbf{RQ2:} Does an architecture designed on \nameDataset transfer to established public bark benchmarks, or is its performance specific to the species and acquisition conditions it was designed around?
\item\textbf{RQ3:} Which architectural components---multi-scale texture aggregation, attention, residual refinement and the colour pathway---account for the observed performance?
\item\textbf{RQ4:} Are the model's decisions grounded in bark texture itself, or in background and contextual correlates that would not survive deployment?
\end{itemize}

\subsection{Contributions}
\label{subsec:contributions}

\begin{itemize}
\item \textbf{A regional bark dataset.} \nameDataset, the first bark image dataset for Bangladesh: 14{,}258 uncropped images of 20 native hardwood and palm species, four districts, three weather conditions, three smartphone models, with English, Bengali and scientific names per class and a fixed stratified split.
\item \textbf{A lightweight texture--colour architecture.} \nameModel, a 2.96\,M-parameter network combining a multi-scale texture pathway with a parallel colour-aware pathway, trained without pre-training, reaching $96.64 \pm 0.66\%$ accuracy over five seeds.
\item \textbf{A controlled efficiency comparison} against nine modern compact CNN and hybrid vision-transformer backbones fine-tuned from ImageNet under an identical protocol, with measured on-device latency and memory (RQ1).
\item \textbf{Cross-dataset evaluation and verification:} the architecture retrained from scratch on BarkVN-50, BarkNet 1.0 and TRUNK12, identifying the regime in which the design fails (RQ2); and Grad-CAM, validated by faithfulness and weight-randomisation checks, confirming that predictions rest on diagnostic bark features (RQ4).
\end{itemize}

RQ3 is answered at the level of the four texture stages by the ablation of Section~\ref{subsec:res-ablation}; finer component-level controls are left to future work.


\section{Background}
\label{sec:background}

\begin{figure*}[!t]
  \centering
  \includegraphics[width=0.6\linewidth]{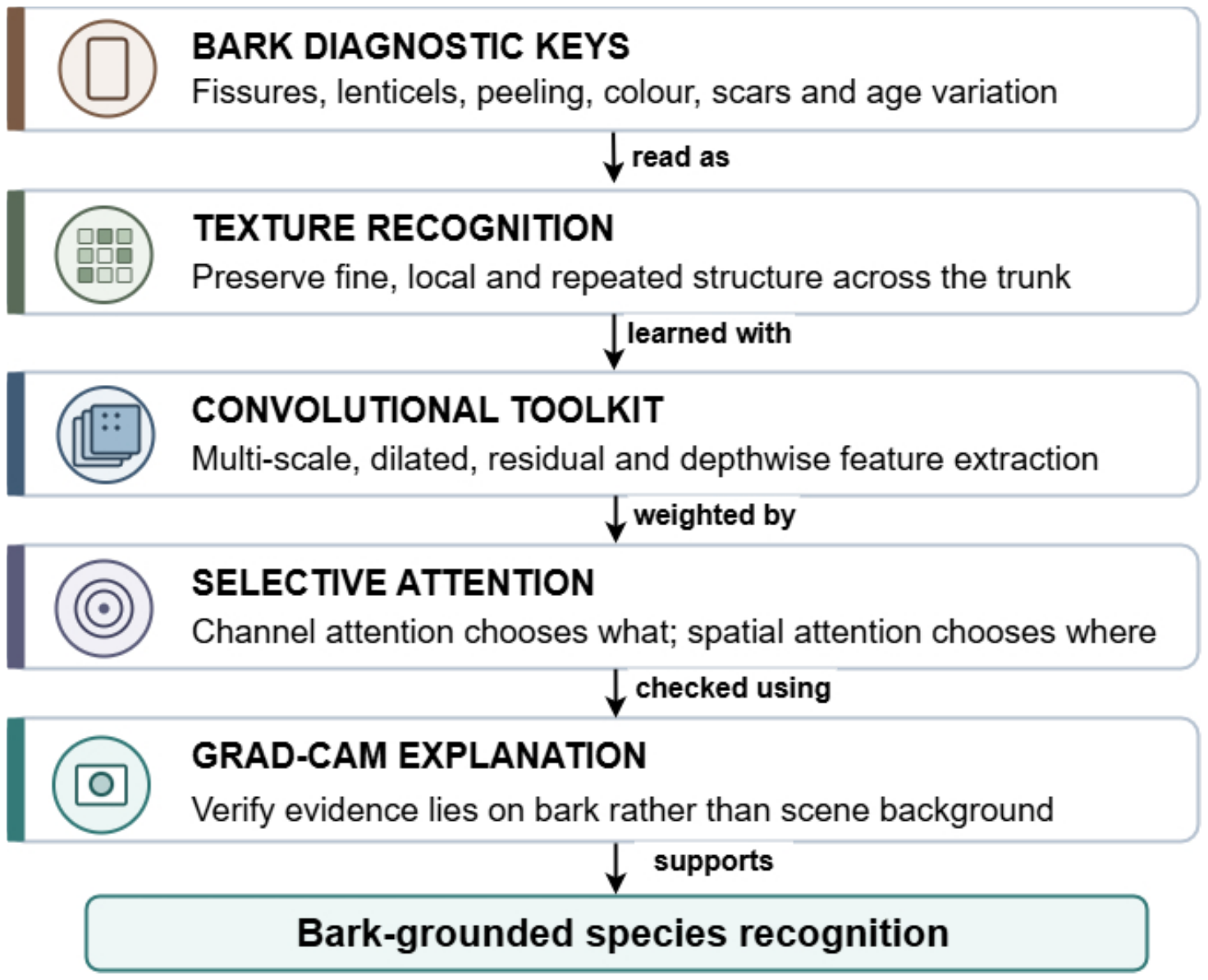}
  \caption{Concepts underlying \nameModel.}
  \label{fig:background}
\end{figure*}

Five ideas underlie the design (Figure~\ref{fig:background}): bark as a diagnostic organ, texture as a recognition problem, the convolutional components we reuse, attention, and post-hoc explanation. Formal definitions are deferred to Section~\ref{sec:method}.

\subsection{Bark as a diagnostic organ}
\label{subsec:bg-bark}

Field botanists read bark from largely textural keys: fissure depth and orientation, lenticel size and arrangement, and exfoliation pattern---whether bark peels in plates, strips, rings or not at all---often decisive at genus level, with colour, resin, lichen and leaf-scar geometry as secondary cues. Two properties shape the learning problem: the keys are local and repetitive, so a classifier must aggregate evidence over the whole field of view rather than localise one object; and they drift with ontogeny, so a corpus of mature individuals need not generalise to saplings. Palms are a special case, carrying persistent leaf-base scars in ring or diamond patterns, distinct from dicot bark though, as we show, not always well separated from one another.

\subsection{Texture recognition, attention and explanation}
\label{subsec:bg-texture}

Texture recognition differs from object recognition in what must be preserved: identity lives in the statistics of fine, repeated structure rather than in coarse shape, so early spatial detail cannot be discarded. Classical descriptors computed those statistics explicitly, interpretable but fixed by the designer; convolutional networks learn the descriptor instead, at the cost of a demand for labelled data that the absence of regional datasets denies. Our design reuses standard devices chosen for the signal: multi-scale Inception aggregation \cite{szegedy2015going,szegedy2016rethinking} and dilated convolution \cite{yu2015multi} to span fissure, lenticel and plate scales within one image; residual bottlenecks \cite{he2016deep} to refine features cheaply; depthwise separable convolution \cite{chollet2017xception,sandler2018mobilenetv2,howard2019searching} to make the colour pathway essentially free; and batch normalisation \cite{ioffe2015batch}, dropout \cite{srivastava2014dropout} and Swish \cite{ramachandran2017searching}. Attention adds a learned decision about which parts of a representation deserve emphasis---channel attention gates each channel by a learned descriptor \cite{hu2018squeeze}, spatial attention chooses where to look \cite{woo2018cbam}---which on uncropped images compensates for the absence of manual cropping.

A classifier accurate for the wrong reason is a field liability: because our images are uncropped and collection is partly site-specific, a network could learn a site's background rather than the trunk, the failure mode documented as context bias \cite{beery2018recognition}. Grad-CAM \cite{selvaraju2017grad} weights the channels of a late feature map by their influence on the predicted class score and sums them into a heat map, so mass on the trunk means the model is using bark and mass on foliage or sky means it is not. Post-hoc and approximate, it answers exactly that question; prior bark studies used the related class-activation-mapping technique \cite{zhou2016learning} for the same purpose \cite{kim2022identifying}, and the wider literature increasingly treats such verification as part of reporting \cite{arrieta2020explainable}.


\section{Related work}
\label{sec:related}

Work on bark-based species identification divides along two axes: how the representation is obtained (handcrafted versus learned) and what is sensed (2D photography versus 3D laser scanning). We review each strand, then the datasets that enabled the learned one, and close by stating the gap this paper addresses. Table~\ref{tab:related} summarises the studies quantitatively.

\subsection{Handcrafted texture descriptors}
\label{subsec:rel-handcrafted}

The earliest systems treated bark identification as classical texture classification, using Gabor filter banks \cite{chi2003plant}, co-occurrence and related statistical features \cite{wan2004bark,song2004bark} and, through the following decade, local binary patterns and their rotation-invariant successors \cite{fekri2020bark,boudra2017statistical,boudra2018plant}, with multispectral Markovian features a notable alternative \cite{remes2019bark}. These descriptors are interpretable, compact and mobile-cheap, and Wan et al.\ already showed that adding colour helps every one of them \cite{wan2004bark}. Their recurring limitation is brittleness away from curated benchmarks: the discriminative pattern is fixed in advance, so it degrades when acquisition conditions shift and cannot adapt to a new species set.

\subsection{Deep learning on bark images}
\label{subsec:rel-deep}

Carpentier et al.~\cite{carpentier2018tree} marked the transition to learned representations with BarkNet 1.0, establishing the lesson we adopt: the number of distinct trees sampled governs generalisation more than raw image count. Subsequent CNN work pushed on scale \cite{faizal2022automated}, on realism (Warner et al.'s uncropped CentralBark \cite{warner2024centralbark}, the regime \nameDataset also operates in) and on efficiency, most closely Wu et al.'s Deep BarkID \cite{wu2021deep}, which reaches deployability by distilling a large pre-trained model rather than designing a compact one. Others targeted specific weaknesses---ROI extraction \cite{ido2019cnn}, the hard 101-species Bark-101 benchmark \cite{ratajczak2019efficient}, patch-based classification \cite{misra2020patch} and retrieval-based re-identification \cite{robert2020tree}. Our earlier work pooled temperate and East Asian corpora with compact interpretable models \cite{ali2025lightweight}; it shares no imagery, no species and no architecture with the present paper, and assembled its network from standard components rather than designing around bark structure. Interpretability entered the field through Kim et al.~\cite{kim2022identifying}, whose class-activation analysis of 42 species found the simpler architecture more generalisable---a motivation for the compact design adopted here.

\subsection{Three-dimensional and multi-organ approaches}
\label{subsec:rel-3d}

A parallel line senses bark geometry from terrestrial laser scans \cite{othmani2013single,othmani2013tree}, whose structural richness comes at an acquisition cost that confines it to research inventories; a further line fuses bark with leaves or other organs \cite{fiel2011automated,bertrand2018bark}, which helps when several organs are visible but not in the off-season case that motivates bark-only recognition.

\subsection{Datasets and the geographic gap}
\label{subsec:rel-datasets}

Public bark corpora span BarkTex \cite{lakmann1998barktex}, TRUNK12 \cite{svab2014trunk12}, AFF \cite{fiel2011automated}, BarkNet 1.0 \cite{carpentier2018tree}, CentralBark \cite{warner2024centralbark}, BarkVN-50 \cite{faizal2022automated} and Bark-101 \cite{ratajczak2019efficient} (Table~\ref{tab:related}), differing in size, species count and, critically for reported accuracy, whether images are cropped to clean bark patches. South Asia entered this list only recently, with BarkVisionAI \cite{chhatre2026barkvisionai}: 13 species from Himachal Pradesh and Odisha under a restrictive CC~BY-NC-ND licence, Himalayan and eastern-Indian rather than deltaic. The Bengal delta is absent altogether: \textit{Mangifera indica}, \textit{Artocarpus heterophyllus}, \textit{Swietenia macrophylla}, \textit{Areca catechu} and \textit{Cocos nucifera} appear in no public bark benchmark and in no pre-trained bark model's label space, and no fine-tuning on existing corpora can supply them.

\subsection{Positioning of this work}
\label{subsec:rel-gap}

Three gaps follow: coverage is narrow and the tropics unrepresented; reported accuracy is entangled with preprocessing, since multi-crop and majority-vote protocols on cropped corpora inflate numbers relative to single-image inference on uncropped ones; and compactness has been treated as post-hoc compression rather than architecture design. This paper addresses all three: an uncropped bark corpus for the Bengal delta; single-image accuracy without multi-crop or voting, across five seeds; and a 2.96\,M-parameter architecture trained from random initialisation, evaluated against nine modern compact backbones under a matched protocol on the accuracy--cost frontier.

\begin{sidewaystable*}[!p]
\centering
\caption{Representative work on bark-based tree species recognition.}
\label{tab:related}
\scriptsize
\setlength{\tabcolsep}{3pt}
\ifdefined\rwunit\else\newlength{\rwunit}\fi\setlength{\rwunit}{0.97\textheight}
\renewcommand{\arraystretch}{0.95}
\begin{tabular}{@{}
>{\raggedright\arraybackslash}p{0.115\rwunit}
>{\raggedright\arraybackslash}p{0.030\rwunit}
>{\raggedright\arraybackslash}p{0.125\rwunit}
>{\raggedright\arraybackslash}p{0.030\rwunit}
>{\raggedright\arraybackslash}p{0.175\rwunit}
>{\raggedright\arraybackslash}p{0.085\rwunit}
>{\raggedright\arraybackslash}p{0.145\rwunit}
>{\raggedright\arraybackslash}p{0.215\rwunit}
@{}}
\toprule
\textbf{Study} & \textbf{Year} & \textbf{Dataset (images)} & \textbf{Cls.} &
\textbf{Method} & \textbf{Region} & \textbf{Protocol / Acc.\ (\%)} &
\textbf{Principal limitation} \\
\midrule
Chi et al.~\cite{chi2003plant} & 2003 & Private (300) & 8 &
Gabor filter banks & --- & single img.\ / 80.0 &
Small curated corpus; no illumination variation. \\
Wan et al.~\cite{wan2004bark} & 2004 & Private (300) & 17 &
RLM / GLCM / HM / ACM with NN, $k$-NN, MMC & --- & single img.\ / 78.0 &
Descriptor comparison only; colour needed for the reported accuracy. \\
Song et al.~\cite{song2004bark} & 2004 & Private (90) & 18 &
Co-occurrence + LCLE on wavelet edges & --- & single img.\ / 88.9 &
90 samples over 18 classes; not reproducible at scale. \\
Fiel \& Sablatnig~\cite{fiel2011automated} & 2011 & AFF (1{,}182) & 11 &
SIFT bag-of-words + GLCM and wavelet features, SVM & Austria &
single img.\ / 69.7 &
Bark far weaker than leaves; corpus not public; SIFT cost precludes online
learning. \\
Othmani et al.~\cite{othmani2013single} & 2013 & TLS, custom & 5 &
3D geometric bark texture, wavelet and contourlet features & France &
single tree / $>$80 &
Requires laser scanning; not field-portable; dataset not public. \\
Othmani et al.~\cite{othmani2013tree} & 2013 & TLS, custom & 5 &
2D deviation maps + roughness and shape, random forest & France &
single tree / 83--100 &
Requires laser scanning; five species; balanced custom set. \\
Othmani et al.~\cite{othmani2016novel} & 2016 & TLS, custom & 6 &
Burst wind segmentation + ScarBook & France & majority vote / 99.6 &
Beech excluded (smooth bark, no scars); dataset not public. \\
Boudra et al.~\cite{boudra2017statistical} & 2017 & BarkTex, AFF, TRUNK12 & 6--12 &
Statistical radial binary patterns & Multi-region & single img.\ / 88.4 &
Beats LBP variants but trails CNNs on unseen data. \\
Carpentier et al.~\cite{carpentier2018tree} & 2018 & BarkNet 1.0 (23{,}616) & 23 &
ResNet-18 / -34, fine-tuned & Canada & single img.\ / 93.9; vote / 97.8 &
Cropped images; Canadian species; voting needs many images per tree. \\
Bertrand et al.~\cite{bertrand2018bark} & 2018 & Pl@ntView (2{,}559) & 101 &
One-vs-all SVM, bark--leaf fusion & France & single img.\ / 30.7 (bark only) &
Bark-only accuracy very low; needs leaves; highly imbalanced. \\
Boudra et al.~\cite{boudra2018plant} & 2018 & BarkTex, TRUNK12, AFF, BarkNet & 6--23 &
Statistical macro binary patterns & Multi-region & single img.\ / 99.8\textsuperscript{*} &
Reported figure not independently confirmed; trails CNN features. \\
Ratajczak et al.~\cite{ratajczak2019efficient} & 2019 &
Bark-101, BarkTex, AFF, TRUNK12 & 6--101 &
Handcrafted descriptors + $k$-NN, SVM & Europe &
single img.\ / 92.4 (BarkTex); 34--42 (Bark-101) &
Bark-101 has shadows, moss, illumination change; collapses at 101 classes. \\
Reme{\v{s}} \& Haindl~\cite{remes2019bark} & 2019 & AFF, BarkTex, TRUNK12 & 6--23 &
Rotationally invariant multispectral Markov features, NN classifier & Multi-region &
single img.\ / 89.1--92.9 &
Handcrafted; limited capacity on large diverse corpora. \\
Ido \& Saitoh~\cite{ido2019cnn} & 2019 & Private & 6 &
ROI extraction + AlexNet / Inception-v3 / VGG & Japan & single img.\ / 74.7 &
Accuracy dominated by ROI quality; six species only. \\
Fekri-Ershad~\cite{fekri2020bark} & 2020 & TRUNK12, AFF & 28 &
Improved local ternary patterns + MLP & Germany, Slovenia, Austria &
single img.\ / 86.8 &
No colour features; hidden-layer count not optimised. \\
Misra et al.~\cite{misra2020patch} & 2020 & Bark-101 (2{,}587) & 101 &
SqueezeNet, MobileNetV2, VGG16, patch-based & Spain & multi-crop / 57.2 &
Low-resolution classes excluded; very low accuracy at 101 classes. \\
Robert et al.~\cite{robert2020tree} & 2020 & Private (2{,}400) & 2 &
DeepBark / SqueezeBark retrieval descriptors & Canada & retrieval / 87.2 (mAP) &
Two species; controlled night-time capture; closed-set only. \\
Wu et al.~\cite{wu2021deep} & 2021 & Indiana Bark (309) + BarkNet & 10 &
MobileNetV2 student, knowledge distillation & USA & multi-crop / 96.1 &
Post-hoc compression of a large pre-trained model; tiny target corpus. \\
Kim et al.~\cite{kim2022identifying} & 2022 & BarkNet 1.0 + custom (9{,}375) & 42 &
VGG-16, EfficientNet, class activation mapping & South Korea & single img.\ / $>$90 &
Confusion within genus and family; deeper model less interpretable. \\
Faizal~\cite{faizal2022automated} & 2022 & BarkVN-50 (5{,}678) & 50 &
ResNet-101, transfer learning & Vietnam & single img.\ / 94.0 &
Large pre-trained backbone; class imbalance; single lighting condition. \\
Cui et al.~\cite{cui2023improvement} & 2023 & BarkNet + BarkNJ (7{,}671) & 33 &
ConvNeXt & China & single img.\ / 97.6 &
Pre-trained weights misidentify congeneric species. \\
Warner et al.~\cite{warner2024centralbark} & 2024 & CentralBark (19{,}147) & 25 &
EfficientNet-B3 / ResNet-50 / MobileNetV3-S & USA & single img.\ / 83.2 &
Uncropped realism costs accuracy; multiple trees per image. \\
Chhatre et al.~\cite{chhatre2026barkvisionai} & 2026 & BarkVisionAI (156k) & 13 &
ResNet, VGG16, EfficientNet, ViT & India & single img.\ / 87.4 &
Two Indian states; degrades in low light; balanced-subset benchmark; restrictive licence. \\
\midrule
\textbf{This work} & \textbf{2026} & \textbf{\nameDataset (14{,}258)} & \textbf{20} &
\textbf{\nameModel, 2.96\,M params, from scratch} & \textbf{Bangladesh} &
\textbf{single img.\ / $\mathbf{96.6 \pm 0.7}$} &
\textbf{First bark corpus for the Bengal delta; no pre-training; uncropped field
imagery; five-seed evaluation.} \\
\bottomrule
\end{tabular}

\vspace{3pt}
\raggedright\scriptsize
\textsuperscript{*}\,Reported by the original authors on a benchmark subset; the figure has not been
independently reproduced. Accuracies in this table are not directly comparable, because the
evaluation protocols (cropped or uncropped imagery, single image, multi-crop or per-tree vote)
differ between studies.
\end{sidewaystable*}


\section{The \nameDataset Dataset}
\label{sec:dataset}

\nameDataset is the first bark image dataset assembled for Bangladesh and, to our knowledge, for the Bengal delta: 14{,}258 photographs of 20 native tree species, captured with consumer smartphones under field conditions and released without cropping or retouching. This section documents the sampling design, acquisition, annotation, composition and fixed split in enough detail to reproduce the collection.

\subsection{Design Rationale}
\label{subsec:data-rationale}

Three decisions shaped the corpus, each trading headline accuracy for external validity. \emph{Images are retained uncropped}, stored as the camera produced them rather than as the hand-isolated trunk patch of several established corpora: cropping raises reported accuracy but removes the commonest failure mode of a deployed tool, a trunk sharing the frame with foliage, stems, litter or sky \cite{warner2024centralbark}. \emph{Sampling is by individual tree}, since the number of distinct individuals governs generalisation more strongly than raw image count \cite{carpentier2018tree}, with a deliberate spread of trunk diameters. And \emph{the class distribution is left unbalanced}, following the relative abundance of these species, because forcing balance would not resemble what a deployed tool encounters.

\subsection{Study Area and Acquisition}
\label{subsec:data-area}

Data were collected in four districts. Rangpur, Dinajpur and Bogura lie in the northern floodplain, where most trees stand on roadsides, in homestead groves and on campuses; Cox's Bazar, on the south-eastern coast, contributes coastal windbreaks and sandy substrate. Four districts were chosen because bark morphology of one species varies with soil, humidity and exposure---a single-site corpus would encode the site rather than the species---and because several species are geographically restricted (\textit{Casuarina equisetifolia} to the coast, \textit{Dillenia indica} and \textit{Syzygium cumini} to the northern groves). Public parks were favoured for their varied illumination and the concentration of species within a walkable area.

Images were captured with three consumer smartphones (Poco M3, Redmi Note 13 Pro, Redmi Note 5 Pro) running Open Camera configured to write unprocessed frames, since disabling vendor-side sharpening, denoising and saturation preserves the fine texture and chromatic values that carry species identity, while three device models introduce realistic sensor and lens variation. For each tree, 10 to 40 photographs were taken from three or four viewpoints and several heights at 20 to 60\,cm, capturing circumferential and vertical variation, and sessions were spread across sunny, cloudy and rainy conditions, which give the same surface visually distinct appearances: hard shadows and specular highlights in sun, flattened relief under cloud, darkened saturated colour in rain.

\subsection{Annotation and Composition}
\label{subsec:data-annotation}

Species identity was determined in the field per individual tree, using leaves, fruit, flowers and habit alongside bark, and verified by an agroforestry specialist (Prof.\ Kazi Kamrul Islam, Department of Agroforestry, Bangladesh Agricultural University). Scientific names follow current accepted nomenclature, with both the accepted binomial and the Bengali name recorded per class (two entries are listed under the currently accepted binomial rather than the more familiar synonym: \textit{Melaleuca viminalis} for \textit{Callistemon viminalis} and \textit{Monoon longifolium} for \textit{Polyalthia longifolia}; a third, \textit{Casuarina equisetifolia}, is recorded with the vernacular ``Jhau'' alongside its English name, since that is how it is known locally).

\begin{figure*}[!t]
  \centering
  \includegraphics[width=\linewidth]{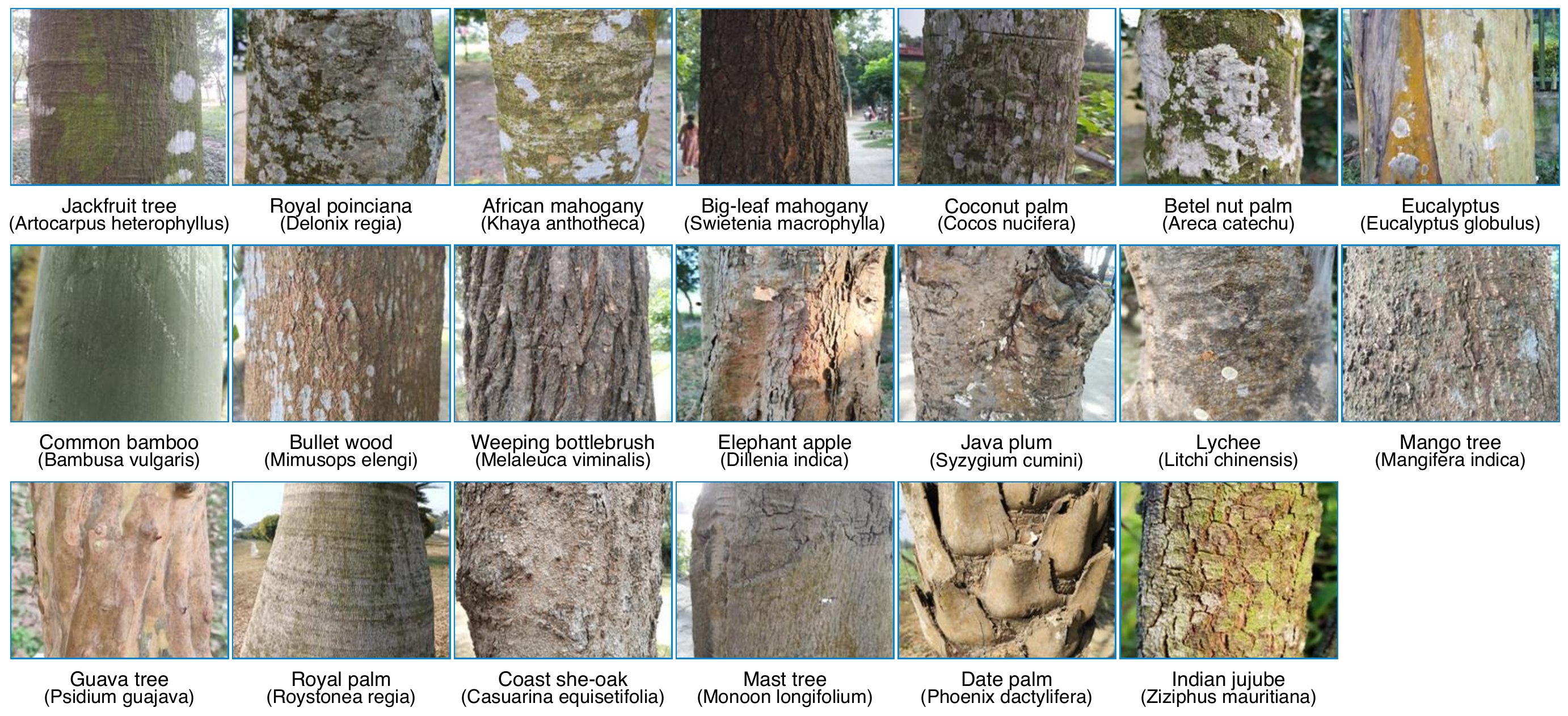}
  \caption{Representative uncropped bark images from the 20 \nameDataset species.}
  \label{fig:dataset_overview}
\end{figure*}

The corpus spans 15 dicot hardwoods, four arecaceous palms (\textit{Cocos nucifera}, \textit{Areca catechu}, \textit{Roystonea regia}, \textit{Phoenix dactylifera}) and one giant grass (\textit{Bambusa vulgaris}); palms and bamboo are included because non-specialists treat them as trees and their leaf-base and internodal sheath scars are a stem regime any recogniser deployed here will meet alongside dicot rhytidome. Per-species tree counts are released with the archive. Table~\ref{tab:tree_class_mapping} lists every class with English, Bengali and scientific names and image counts; Figure~\ref{fig:dataset_overview} shows representative patches. Counts range from 1{,}389 for \textit{Swietenia macrophylla} to 249 for \textit{Casuarina equisetifolia}, an imbalance of 5.6 to 1.

\begin{table*}[!t]
\centering
\caption{The 20 \nameDataset species, with English, Bengali and scientific names and image counts.}
\label{tab:tree_class_mapping}
\small
\renewcommand{\arraystretch}{1.15}
\begin{tabularx}{\textwidth}{@{}l X l r@{}}
\toprule
\textbf{Class Name} & \textbf{Real Name} & \textbf{Scientific Name} & \textbf{No. of Samples} \\
\midrule
Class 0 & Jackfruit tree (\bn{কাঁঠাল গাছ}) & \textit{Artocarpus heterophyllus} & 853 \\
Class 1 & Royal poinciana / Flamboyant (\bn{কৃষ্ণচূড়া}) & \textit{Delonix regia} & 908 \\
Class 2 & African mahogany (\bn{আফ্রিকান মেহগনি}) & \textit{Khaya anthotheca} & 1085 \\
Class 3 & Big-leaf mahogany (\bn{বড় পাতার মেহগনি}) & \textit{Swietenia macrophylla} & 1389 \\
Class 4 & Coconut palm (\bn{নারকেল গাছ}) & \textit{Cocos nucifera} & 871 \\
Class 5 & Betel nut palm (\bn{সুপারি গাছ}) & \textit{Areca catechu} & 943 \\
Class 6 & Eucalyptus (\bn{ইউক্যালিপটাস}) & \textit{Eucalyptus globulus} & 588 \\
Class 7 & Common bamboo (\bn{বাঁশ}) & \textit{Bambusa vulgaris} & 375 \\
Class 8 & Bullet wood (\bn{বকুল}) & \textit{Mimusops elengi} & 833 \\
Class 9 & Weeping bottlebrush (\bn{বোতলব্রাশ}) & \textit{Melaleuca viminalis} & 587 \\
Class 10 & Elephant apple (\bn{চালতা}) & \textit{Dillenia indica} & 729 \\
Class 11 & Java plum / Indian blackberry (\bn{জাম}) & \textit{Syzygium cumini} & 600 \\
Class 12 & Lychee (\bn{লিচু}) & \textit{Litchi chinensis} & 687 \\
Class 13 & Mango tree (\bn{আম গাছ}) & \textit{Mangifera indica} & 584 \\
Class 14 & Guava tree (\bn{পেয়ারা গাছ}) & \textit{Psidium guajava} & 542 \\
Class 15 & Royal palm (\bn{রয়্যাল পাম}) & \textit{Roystonea regia} & 612 \\
Class 16 & Coast she-oak / Jhau (\bn{ঝাউ গাছ}) & \textit{Casuarina equisetifolia} & 249 \\
Class 17 & Mast tree / False Ashoka (\bn{দেবদারু}) & \textit{Monoon longifolium} & 725 \\
Class 18 & Date palm (\bn{খেজুর গাছ}) & \textit{Phoenix dactylifera} & 493 \\
Class 19 & Indian jujube (\bn{বরই}) & \textit{Ziziphus mauritiana} & 605 \\
\midrule
\multicolumn{3}{@{}l}{\textbf{Total}} & \textbf{14{,}258} \\
\bottomrule
\end{tabularx}
\end{table*}

\subsection{Preparation Pipeline and Splits}
\label{subsec:data-pipeline}

\begin{figure*}[!t]
  \centering
  \includegraphics[width=0.8\linewidth]{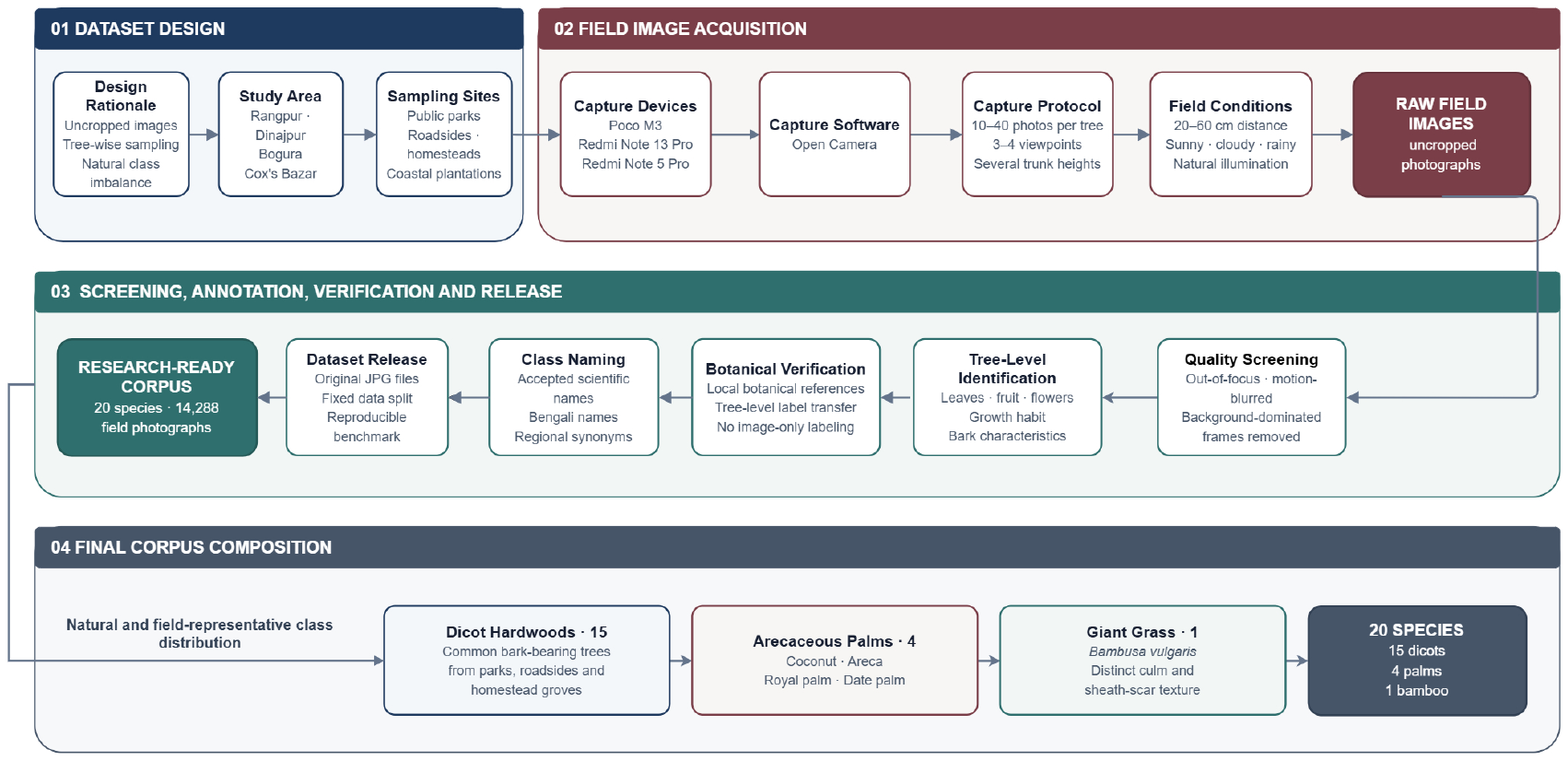}
  \caption{The \nameDataset preparation pipeline.}
  \label{fig:pipeline}
\end{figure*}

Figure~\ref{fig:pipeline} summarises the path from field capture to the released archive. Raw frames pass a quality screen removing images out of focus, severely motion-blurred or dominated by background; survivors inherit the verified label of their source tree and are filed by class, with no geometric or photometric editing and resizing deferred to the model input pipeline (Section~\ref{subsec:preproc}). The corpus is partitioned once, stratified by class at the image level, into 70/15/15 subsets---9{,}971 training, 2{,}138 validation and 2{,}149 test images---so the proportions of Table~\ref{tab:tree_class_mapping} are preserved in every subset and no image appears in more than one. The partition is fixed and shipped with the dataset, and every number in this paper refers to it, allowing subsequent work to report directly comparable figures.

One consequence must be stated plainly. Because each tree contributes 10 to 40 frames, an image-level partition places different photographs of the same trunk on both sides of the split. The accuracies reported here therefore measure recognition of the sampled individuals under unseen viewpoints, distances and illumination, not generalisation to individuals never seen in training, and they are an upper bound on the latter (Section~\ref{subsec:disc-limitations}). We keep the image-level partition as the released benchmark because it is the one every number in this paper refers to, and the archive records the source tree of each image, so a tree-grouped partition can be constructed from it without further collection.

\subsection{Relation to Existing Bark Corpora}
\label{subsec:data-relation}

\nameDataset is an order of magnitude larger than BarkTex \cite{lakmann1998barktex} and TRUNK12 \cite{svab2014trunk12}, comparable in size to CentralBark \cite{warner2024centralbark} and BarkNet 1.0 \cite{carpentier2018tree} and, like CentralBark, published uncropped, while its species set overlaps none of them and none of the 13 Himalayan and eastern-Indian species of BarkVisionAI \cite{chhatre2026barkvisionai}. The taxa dominating the Bengal delta remain absent from every public bark benchmark, and that absence is what \nameDataset addresses.


\section{Proposed Method}
\label{sec:method}

This section specifies the architecture, the optimisation recipe and the evaluation protocol. Every symbol is collected in Supplementary Table~S1.

\subsection{Problem Formulation}
\label{subsec:problem}

Given a labelled corpus of RGB bark photographs over $C = 20$ species, we seek $f_{\boldsymbol{\theta}}$ mapping an image to a class-probability vector, with $\hat{y} = \argmax_{c} [f_{\boldsymbol{\theta}}(\mathbf{x})]_c$. Two constraints separate this from generic classification: $\boldsymbol{\theta}$ is small enough for on-device inference and is initialised at random rather than from an ImageNet checkpoint, so all discriminative structure is learned from the corpus alone.

\subsection{Preprocessing and Augmentation}
\label{subsec:preproc}

Every image is resized to $224 \times 224$ and scaled to $[0,1]$; no mean subtraction or standardisation is applied, since the colour pathway of Section~\ref{subsubsec:cam} operates on chromatic content that standardisation would partly remove. Training images pass a stochastic operator of independent horizontal and vertical flips (bark has no canonical orientation), rotation in $[-0.2\pi, 0.2\pi]$, zoom in $[0.8, 1.2]$ and $\pm 20\%$ contrast and brightness jitter mirroring the sampled weather; augmentation runs after the cache so each epoch sees a fresh realisation, and validation and test data are never augmented. A second stage applies, per batch with equal probability, one of MixUp \cite{zhang2018mixup} ($\gamma \sim \mathrm{Beta}(0.2,0.2)$) or CutMix \cite{yun2019cutmix} (target weighted by retained area): MixUp discourages over-confidence on a single texture prototype and CutMix forces evidence to accumulate over the whole field of view; both give soft targets, so the label-smoothing factor of Section~\ref{subsec:optimisation} is halved.

\subsection{Architecture of \nameModel}
\label{subsec:architecture}

\begin{figure*}[!t]
  \centering
  \includegraphics[width=\textwidth]{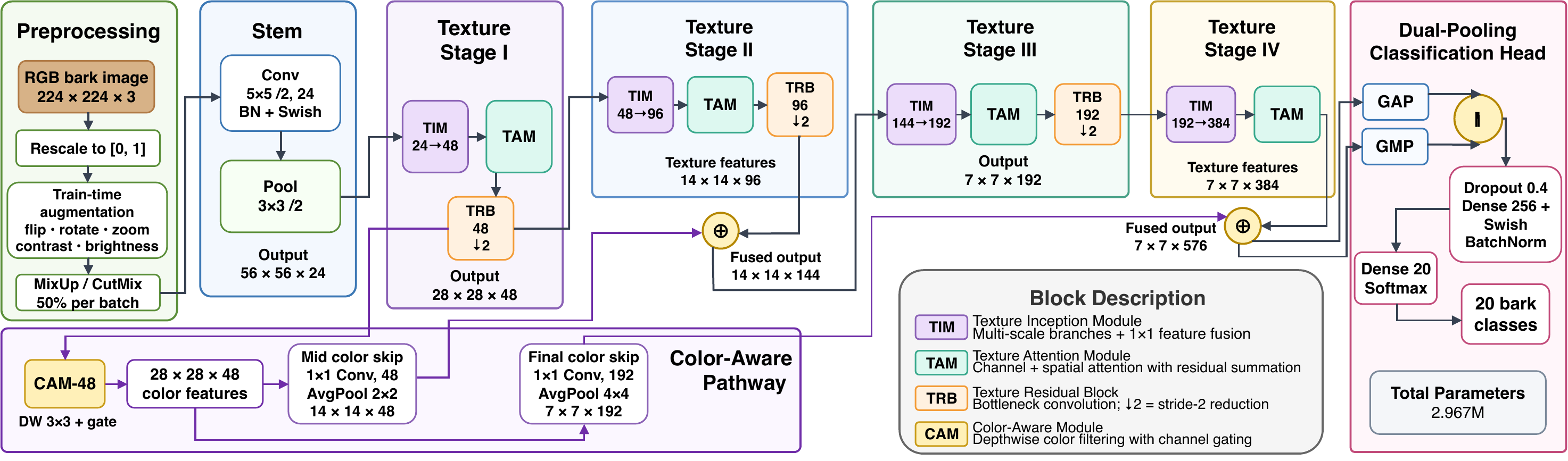}
  \caption{The \nameModel architecture: texture pathway (top), colour pathway (bottom) and dual-pooling head.}
  \label{fig:arch}
\end{figure*}

\nameModel consists of a stem, four texture stages, a parallel colour pathway with two fusion points, and a dual-pooling head. Figure~\ref{fig:arch} gives the block diagram with tensor shapes and Table~\ref{tab:arch} the stage-by-stage configuration. The texture pathway grows in receptive field while a colour pathway tapped early is reinjected later, so chromatic evidence reaches the classifier without surviving the full texture hierarchy.

\subsubsection{Stem}
\label{subsubsec:stem}

The stem reduces resolution fourfold via a stride-2 $5 \times 5$ convolution with batch normalisation \cite{ioffe2015batch} and Swish \cite{ramachandran2017searching}, then stride-2 max-pooling, yielding a $56 \times 56 \times 24$ tensor. A $5 \times 5$ kernel is preferred over $3 \times 3$ because the first layer must span a fissure width to respond to it.

\subsubsection{Texture Inception Module}
\label{subsubsec:tim}

The \tim aggregates evidence at six receptive-field configurations in parallel. For input $\mathbf{Z}$ and width $F$, with $\phi(\cdot) = \mathrm{ReLU}(\BN(\cdot))$, the branches are
\begin{align}
  \mathbf{B}_1 &= \phi(\mathbf{W}^{(1)}_{1 \times 1} * \mathbf{Z}), \label{eq:tim1}\\
  \mathbf{B}_2 &= \phi(\mathbf{W}^{(2b)}_{3 \times 3} * \phi(\mathbf{W}^{(2a)}_{1 \times 1} * \mathbf{Z})), \label{eq:tim2}\\
  \mathbf{B}_3 &= \phi(\mathbf{W}^{(3c)}_{3 \times 3} * \phi(\mathbf{W}^{(3b)}_{3 \times 3} * \phi(\mathbf{W}^{(3a)}_{1 \times 1} * \mathbf{Z}))), \label{eq:tim3}\\
  \mathbf{B}_4 &= \phi(\mathbf{W}^{(4)}_{3 \times 3} *_{d=2} \mathbf{Z}), \label{eq:tim4}\\
  \mathbf{B}_5 &= \phi(\mathbf{W}^{(5)}_{3 \times 3} *_{d=4} \mathbf{Z}), \label{eq:tim5}\\
  \mathbf{B}_6 &= \phi(\mathbf{W}^{(6)}_{1 \times 5} * \mathbf{Z}), \label{eq:tim6}
\end{align}
where $*_{d}$ is dilated convolution with rate $d$. Every branch emits $F$ channels; the six are concatenated and projected by a $1 \times 1$ convolution back to $2F$,
\begin{equation}
  \tim(\mathbf{Z}; F) = \phi\big( \mathbf{W}_{1 \times 1} * \Concat(\mathbf{B}_1, \dots, \mathbf{B}_6) \big) \in \R^{H \times W \times 2F}.
  \label{eq:tim}
\end{equation}
Each branch targets a bark structure: $\mathbf{B}_1$ mixes channels, $\mathbf{B}_2$ captures lenticel-scale detail, $\mathbf{B}_3$ is the factorised $5 \times 5$ of Szegedy et al.~\cite{szegedy2016rethinking} for plate-scale structure, $\mathbf{B}_4$ and $\mathbf{B}_5$ use dilation \cite{yu2015multi} to span $5 \times 5$ and $9 \times 9$ input windows at the cost of a $3 \times 3$ kernel and without downsampling, and the anisotropic $1 \times 5$ $\mathbf{B}_6$ responds to the elongated leaf-base scars of palms. Because bark has no canonical orientation, the flip and rotation augmentation of Section~\ref{subsec:preproc} supplies the transposed response that a single $1 \times 5$ branch does not.

\subsubsection{Texture Attention Module}
\label{subsubsec:tam}

The \tam applies channel and spatial gating in parallel and merges both with the identity. The channel branch is squeeze-and-excitation \cite{hu2018squeeze} with reduction $r = 8$, producing $\mathbf{Z}^{\mathrm{ch}}$; the spatial branch pools across channels by average and max, learns a single-channel gate from a $5 \times 5$ convolution over the two maps, and gives $\mathbf{Z}^{\mathrm{sp}}$. The module returns the additive combination
\begin{equation}
  \tam(\mathbf{Z}) = \mathbf{Z}^{\mathrm{ch}} + \mathbf{Z}^{\mathrm{sp}} + \mathbf{Z}.
  \label{eq:tam}
\end{equation}
The additive form is deliberate: sequential multiplication as in CBAM \cite{woo2018cbam} suppresses outright a region rejected by either gate, whereas adding lets strong channel evidence survive weak spatial evidence and the reverse, which is safer on uncropped images; the retained identity lets the module degenerate to a no-op if gating proves unhelpful.

\subsubsection{Texture Residual Block}
\label{subsubsec:trb}

The \trb is a standard bottleneck residual unit that also halves resolution: a $1 \times 1$ stride-2 reduction to a quarter of the output width, a $3 \times 3$ convolution, a $1 \times 1$ expansion and a projection shortcut, all with batch normalisation and ReLU. The fourfold bottleneck runs the only expensive $3 \times 3$ convolution at a quarter width, which holds the parameter budget at 2.96\,M despite the width of the \tim.

\subsubsection{Colour-Aware Module and Colour Skips}
\label{subsubsec:cam}

Repeated stride-2 reduction destroys chromatic detail before the classifier sees it, yet colour is a genuine, if secondary, discriminant. The \cam branches off after the first stage, where chromatic information is still largely intact, and carries it on its own track through a pointwise, depthwise \cite{chollet2017xception} and pointwise factorisation followed by a squeeze-and-excitation gate $\mathbf{g}$,
\begin{equation}
  \cam(\mathbf{Z}; F)_{ijc} = g_c \, [\mathbf{V}_3]_{ijc},
  \label{eq:cam}
\end{equation}
where $\mathbf{V}_3$ is the factorised output. This keeps the pathway under 4\% of the parameters. The colour tensor is reinjected at two depths---after texture stage~II ($14 \times 14$) and immediately before the head ($7 \times 7$)---each time projected by a $1 \times 1$ convolution and average-pooled to match resolution, average pooling preserving mean chromaticity (Table~\ref{tab:arch}).

\subsubsection{Dual-Pooling Head}
\label{subsubsec:head}

Texture identity is carried both by a detector's average response over the trunk (dense lenticels) and by its strongest single response (a few deep fissures), which are not interchangeable, so the head concatenates global average and global max pooling of the final fused tensor, applies dropout $0.4$ \cite{srivastava2014dropout}, and passes a 256-dimensional Swish embedding with batch normalisation to the softmax classifier; every kernel carries an $\ell_2$ penalty $\lambda = 10^{-5}$.

\subsubsection{Complexity}
\label{subsubsec:complexity}

The model holds 2{,}957{,}956 trainable parameters (2{,}967{,}228 total) and occupies 11.32\,MB on disk (Table~\ref{tab:arch}), one to two orders of magnitude below the backbones typically fine-tuned for this task.

\begin{table}[!t]
\centering
\caption{Stage-by-stage configuration of \nameModel ($\downarrow 2$: stride-2 reduction).}
\label{tab:arch}
\scriptsize
\setlength{\tabcolsep}{4pt}
\renewcommand{\arraystretch}{1.15}
\begin{tabular}{@{}llc@{}}
\toprule
\textbf{Stage} & \textbf{Composition} & \textbf{Output shape} \\
\midrule
Input & Rescale $1/255$, augmentation & $224 \times 224 \times 3$ \\
\addlinespace[2pt]
Stem & $5 \times 5$ conv-24 $\downarrow 2$, BN, Swish, & $56 \times 56 \times 24$ \\
     & $3 \times 3$ max-pool $\downarrow 2$ & \\
\addlinespace[2pt]
Texture I & \tim(24), \tam, \trb(48, $\downarrow 2$) & $28 \times 28 \times 48$ \\
Colour tap & \cam(48) & $28 \times 28 \times 48$ \\
\addlinespace[2pt]
Texture II & \tim(48), \tam, \trb(96, $\downarrow 2$) & $14 \times 14 \times 96$ \\
Mid fusion & Concat with $1 \times 1$ conv-48, & $14 \times 14 \times 144$ \\
           & avg-pool $2 \times 2$ & \\
\addlinespace[2pt]
Texture III & \tim(96), \tam, \trb(192, $\downarrow 2$) & $7 \times 7 \times 192$ \\
Texture IV & \tim(192), \tam & $7 \times 7 \times 384$ \\
Final fusion & Concat with $1 \times 1$ conv-192, & $7 \times 7 \times 576$ \\
             & avg-pool $4 \times 4$ & \\
\addlinespace[2pt]
Head & $\GAP \, \| \, \GMP$, dropout 0.4, & $256$ \\
     & dense-256, Swish, BN & \\
Classifier & Dense-20, softmax & $20$ \\
\midrule
\multicolumn{2}{@{}l}{\textbf{Trainable parameters}} & \textbf{2{,}957{,}956} \\
\multicolumn{2}{@{}l}{\textbf{Total parameters}} & \textbf{2{,}967{,}228} \\
\multicolumn{2}{@{}l}{\textbf{Model size}} & \textbf{11.32\,MB} \\
\bottomrule
\end{tabular}
\end{table}

\subsection{Optimisation}
\label{subsec:optimisation}

The network minimises categorical cross-entropy with label smoothing \cite{muller2019does} at $\varepsilon = 0.05$ plus the $\ell_2$ penalty; smoothing matters because several species are genuinely ambiguous from a single patch, and the value is halved because batch mixing already supplies soft targets. Parameters are updated with Adam \cite{kingma2015adam} at base rate $10^{-3}$, batch size 32, under a 5-epoch linear warmup then cosine decay \cite{loshchilov2017sgdr} to $10^{-6}$ over 300 epochs; warmup avoids the large early updates that destabilise batch-normalisation statistics in a from-scratch network. Training halts when validation accuracy stalls for 40 epochs, restoring the best-validation weights.

\subsection{Experimental Protocol}
\label{subsec:protocol}

Six experiments answer the four research questions.

\textbf{E1: Benchmark on \nameDataset.} \nameModel is trained, selected on validation and evaluated once on test, over five seeds to separate architecture from initialisation noise; we report mean and standard deviation, with a detailed analysis of the reference run. Inference is single-image throughout---no multi-crop averaging, test-time augmentation or per-tree aggregation.

\textbf{E2: Comparison with modern backbones.} \nameModel is compared against nine recent CNN and hybrid vision-transformer backbones initialised from ImageNet and fine-tuned on \nameDataset: MobileNetV4 Conv-Small and Hybrid-Medium \cite{qin2024mobilenetv4}, EfficientNetV2-B0 and -S \cite{tan2021efficientnetv2}, EfficientFormerV2-S0 and -S1 \cite{li2023rethinking}, MobileViT-S \cite{mehta2022mobilevit}, MobileViTv2-100 \cite{mehta2023separable} and GhostNetV2-100 \cite{tang2022ghostnetv2}. Every baseline uses the identical split, preprocessing, augmentation, optimiser, callbacks and epoch budget, so architecture and initialisation are the only variables; the comparison therefore bounds the joint cost of designing a compact network \emph{and} forgoing pre-training, and does not isolate either. Each baseline is trained once, so the baseline figures carry no run-to-run interval and must be read against the 0.66-point seed spread we measure for \nameModel. Parameters and MACs are counted analytically over one $224 \times 224$ forward pass.

\textbf{E3: Deployment behaviour.} \nameModel is exported to TensorFlow Lite in three forms---single precision, dynamic-range weight quantisation, and full-integer quantisation calibrated on training images---each evaluated on test before timing. On-device timing uses the TensorFlow Lite benchmark tool over seven execution configurations (reference kernels, XNNPACK, NNAPI and the GPU delegate at one to four threads), each repeated three times with $\ge$100 timed inferences after 20 warmup iterations; we report mean and standard deviation, with host latency for reference.

\textbf{E4: Cross-dataset generalisation.} To answer RQ2, the architecture (not the trained weights) is retrained from scratch, since label spaces differ, on BarkNet 1.0 \cite{carpentier2018tree}, BarkVN-50 \cite{faizal2022automated} and TRUNK12 \cite{svab2014trunk12}, each using its published split where one exists and a stratified 70/15/15 split otherwise, with only the classifier width changed.

\textbf{E5: Ablation study.} To answer RQ3 we report a stage-level ablation, removing one whole texture stage at a time and retraining from scratch, reading contribution against parameter count so a stage's effect is not confounded with capacity. Every variant is trained for a fixed 100 epochs, whereas the full model trains to a validation plateau at a mean best epoch of 216; the variants are therefore budget-matched to one another but not to the full model, so the reported drops bound each removal from above and it is the ordering, not the absolute point deltas, that the analysis rests on. Because the run-to-run standard deviation is 0.66 points, the single-seed removals are decisive only for the three deeper ones (margins $\ge 1.3$ points, confirmed by McNemar's test \cite{mcnemar1947note}), while the shallowest (0.08 points) is reported as unresolved. Finer component-level controls (isolating the \tim, \tam, \trb and each colour skip) are left to future work.

\textbf{E6: Explainability.} To answer RQ4 we apply Grad-CAM \cite{selvaraju2017grad} and Grad-CAM++ \cite{chattopadhay2018grad} to the last convolutional layer of the final texture stage, validating the maps on 200 test images with the deletion and insertion measures of Petsiuk et al.~\cite{petsiuk2018rise} against a random-attribution control and the cascading weight-randomisation sanity check of Adebayo et al.~\cite{adebayo2018sanity}.

\subsection{Evaluation Metrics}
\label{subsec:metrics}

We report accuracy and per-class precision, recall and F1, aggregated as an unweighted macro mean and weighted by support (the weighted mean describes field performance under the natural distribution, the macro mean weights a rare species as heavily as a common one), together with top-$k$ accuracy, since a field tool may present a shortlist, and expected calibration error over 15 equal-width bins \cite{guo2017calibration}. Weighted recall coincides with accuracy by construction and is not tabulated separately.

\subsection{Implementation and Reproducibility}
\label{subsec:implementation}

Models are implemented in TensorFlow/Keras and trained on GPU. The input pipeline caches decoded images before augmentation and prefetches asynchronously, with batch mixing applied after augmentation so both stages see fresh randomness each epoch. The multi-seed study of Section~\ref{subsec:res-multiseed} sweeps seeds 42 to 46. The training curves, confusion matrix, per-species breakdown, attributions and exported artefacts of Sections~\ref{subsec:res-overall}--\ref{subsec:res-calibration} are all computed from one further training run under the identical recipe, split and hyper-parameters, referred to throughout as the \emph{reference run}; reporting them from a single checkpoint keeps the error analysis, the explanations and the deployment measurements mutually consistent, and its aggregate accuracy falls inside the seed range of Table~\ref{tab:multiseed} (Section~\ref{subsec:res-overall}).


\section{Results}
\label{sec:results}

Unless stated otherwise, every number comes from single-image inference on the held-out \nameDataset test split ($n = 2{,}149$): no multi-crop averaging, test-time augmentation or per-tree aggregation. This is stricter than several studies in Table~\ref{tab:related}, which should be kept in mind when comparing absolute values. The split is stratified by image rather than by tree (Section~\ref{subsec:data-pipeline}), so these figures measure recognition of the sampled individuals under unseen viewpoints and illumination, and are an upper bound on generalisation to unseen individuals.

\subsection{Stability across seeds}
\label{subsec:res-multiseed}

Trained five times from independent initialisations (seeds 42--46) under an identical recipe and evaluated on the same test split (Table~\ref{tab:multiseed}), the architecture is stable: test accuracy $96.64 \pm 0.66\%$ and weighted F1 $96.61 \pm 0.68\%$, a spread of 1.58 points between weakest and strongest seed (95.77--97.35\%). Weighted recall equals accuracy by construction; weighted precision and F1 track it within 0.25 points, as expected when the model does not exploit the class imbalance. Run length varies more than performance: early stopping fires on a seed-dependent validation plateau, and because training continues through the 40-epoch patience window beyond the best epoch listed in Table~\ref{tab:multiseed}, the five runs occupy $256 \pm 47$ epochs and $239 \pm 48$ minutes each on one GPU. We therefore report the multi-seed mean as the headline result, and the 0.66-point standard deviation sets the resolution of every later comparison: differences below roughly 1.3 points between single runs should not be interpreted.

\begin{table}[!t]
\centering
\caption{Per-seed performance of \nameModel on \nameDataset (five seeds).}
\label{tab:multiseed}
\footnotesize
\setlength{\tabcolsep}{4pt}
\renewcommand{\arraystretch}{1.12}
\begin{tabular}{@{}c r r r r r r@{}}
\toprule
\textbf{Seed} & \textbf{Best} & \textbf{Val.} & \textbf{Acc.} &
\textbf{Prec.} & \textbf{F1} & \textbf{Time} \\
 & \textbf{epoch} & \textbf{(\%)} & \textbf{(\%)} & \textbf{(\%)} &
\textbf{(\%)} & \textbf{(min)} \\
\midrule
42 & 196 & 97.19 & 96.65 & 96.77 & 96.66 & 216 \\
43 & 143 & 96.96 & 95.77 & 95.89 & 95.72 & 167 \\
44 & 233 & 97.85 & 97.21 & 97.30 & 97.20 & 251 \\
45 & 255 & 97.75 & 97.35 & 97.42 & 97.32 & 276 \\
46 & 253 & 97.33 & 96.23 & 96.44 & 96.17 & 284 \\
\midrule
\textbf{Mean} & \textbf{216} & \textbf{97.42} & \textbf{96.64} &
\textbf{96.77} & \textbf{96.61} & \textbf{239} \\
\textbf{S.d.} & \textbf{47} & \textbf{0.38} & \textbf{0.66} &
\textbf{0.63} & \textbf{0.68} & \textbf{48} \\
\bottomrule
\end{tabular}
\end{table}

\subsection{The reference run}
\label{subsec:res-overall}

The remainder of this section analyses the reference run of Section~\ref{subsec:implementation}: one further training run under the same recipe, split and hyper-parameters, whose training history and single checkpoint supply the training curves, the confusion matrix, the per-species breakdown, the attribution maps and the exported artefacts, so that they describe one model rather than an average over five. It attains 95.91\% accuracy, 96.10\% weighted precision and 95.88\% weighted F1, with 88 of 2{,}149 test images misclassified. That falls between the weakest and the second-weakest seed of Table~\ref{tab:multiseed} and 0.73 points below the five-seed mean, so every number in the rest of this section is a conservative rather than a favourable reading of the architecture. Macro and weighted averages differ by at most 0.31 points despite the 5.6:1 imbalance (macro recall 96.22\% slightly exceeds weighted recall 95.91\%), so the model is not trading rare-class for common-class accuracy.

\subsection{Training behaviour}
\label{subsec:res-training}

\begin{figure}[!t]
  \centering
  \includegraphics[width=\linewidth]{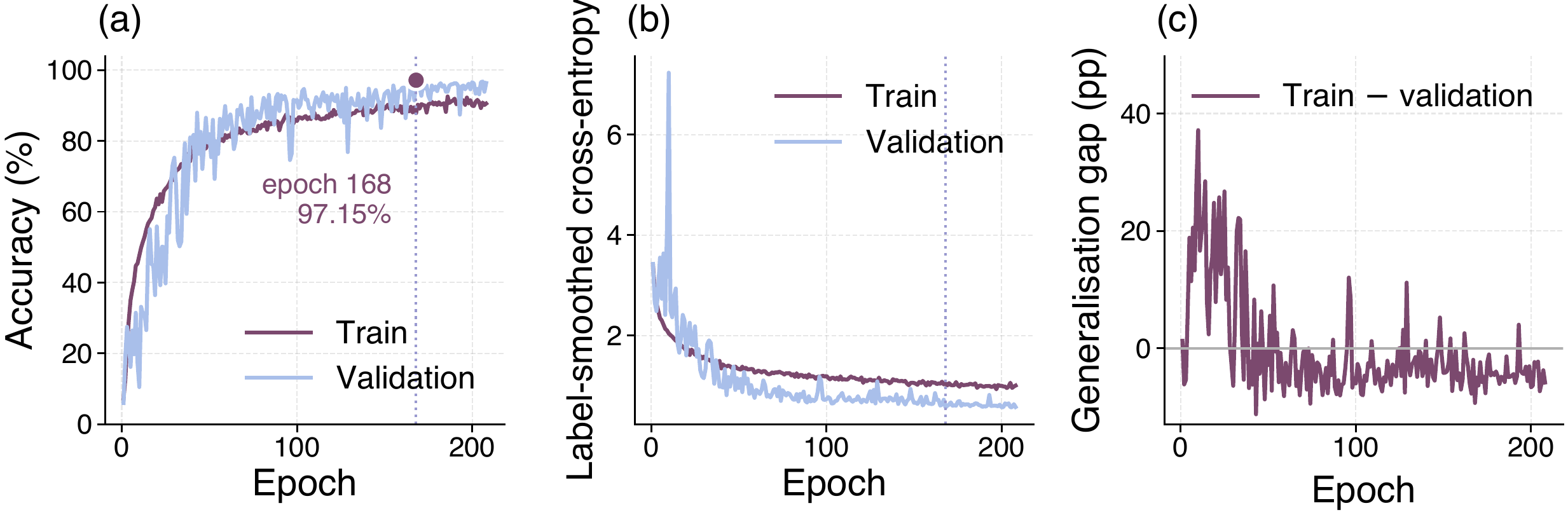}
  \caption{Training dynamics of the \nameModel reference run on \nameDataset: (a) accuracy, (b) loss, (c) generalisation gap.}
  \label{fig:training}
\end{figure}

Figure~\ref{fig:training} shows the accuracy, loss and generalisation-gap traces of that same run. The validation trace is noisier than the training trace, since augmentation applies only to training batches; the generalisation gap oscillates early then contracts as the cosine schedule anneals the learning rate, and the two losses descend together without diverging, so dropout, the $\ell_2$ penalty, label smoothing and augmentation together prevent overfitting at this capacity.

\subsection{Per-species performance}
\label{subsec:res-perclass}

Per-species results appear in Table~\ref{tab:perclass}. Sixteen of twenty species exceed 95\% F1 and eleven exceed 97\%; \textit{Phoenix dactylifera} is solved outright (100.00\% F1), followed by \textit{Ziziphus mauritiana}, \textit{Eucalyptus globulus} and \textit{Mimusops elengi} (99.19--99.45\%), each with a highly regular surface. Weakness concentrates in four classes: \textit{Litchi chinensis} is hardest at 85.25\% F1 (recall 75.00\% against precision 98.73\%---rarely predicted wrongly, but a quarter of its images assigned elsewhere); \textit{Mangifera indica} shows the mirror pattern (precision 85.42\%, recall 93.18\%), as does \textit{Monoon longifolium} (precision 85.48\%, recall 97.25\%), so both act as attractors while lychee is the principal donor; \textit{Delonix regia} is the fourth (91.43\% F1, precision and recall balanced). Per-species F1 is uncorrelated with class size ($r = -0.12$; Supplementary Figure~S1b): \textit{Casuarina equisetifolia}, the smallest class at 249 images, reaches 97.44\% F1 with perfect recall, while \textit{Litchi chinensis} (687 images) records the worst score. What limits performance is intrinsic bark distinctiveness, not image count, so additional effort should target acquisition protocol---viewpoint, standoff, trunk maturity---rather than raw volume.

\begin{table}[!t]
\centering
\caption{Per-species performance of the \nameModel reference run, ordered by F1.}
\label{tab:perclass}
\footnotesize
\setlength{\tabcolsep}{4pt}
\renewcommand{\arraystretch}{1.06}
\begin{tabular}{@{}c l r r r r@{}}
\toprule
\textbf{Cls.} & \textbf{Species} & \textbf{Prec.} & \textbf{Rec.} &
\textbf{F1} & \textbf{Sup.} \\
\midrule
18 & \textit{Phoenix dactylifera}      & 100.00 & 100.00 & 100.00 & 74 \\
19 & \textit{Ziziphus mauritiana}      & 100.00 & 98.90 & 99.45 & 91 \\
6  & \textit{Eucalyptus globulus}      & 100.00 & 98.88 & 99.44 & 89 \\
8  & \textit{Mimusops elengi}          & 100.00 & 98.40 & 99.19 & 125 \\
5  & \textit{Areca catechu}            & 99.29 & 98.59 & 98.94 & 142 \\
15 & \textit{Roystonea regia}          & 98.90 & 97.83 & 98.36 & 92 \\
14 & \textit{Psidium guajava}          & 96.47 & 100.00 & 98.20 & 82 \\
9  & \textit{Melaleuca viminalis}      & 98.85 & 96.63 & 97.73 & 89 \\
16 & \textit{Casuarina equisetifolia}  & 95.00 & 100.00 & 97.44 & 38 \\
7  & \textit{Bambusa vulgaris}         & 96.55 & 98.25 & 97.39 & 57 \\
10 & \textit{Dillenia indica}          & 96.43 & 98.18 & 97.30 & 110 \\
4  & \textit{Cocos nucifera}           & 96.92 & 96.18 & 96.55 & 131 \\
0  & \textit{Artocarpus heterophyllus} & 96.85 & 96.09 & 96.47 & 128 \\
2  & \textit{Khaya anthotheca}         & 94.12 & 98.16 & 96.10 & 163 \\
3  & \textit{Swietenia macrophylla}    & 98.00 & 93.78 & 95.84 & 209 \\
11 & \textit{Syzygium cumini}          & 95.60 & 95.60 & 95.60 & 91 \\
\midrule
1  & \textit{Delonix regia}            & 89.51 & 93.43 & 91.43 & 137 \\
17 & \textit{Monoon longifolium}       & 85.48 & 97.25 & 90.99 & 109 \\
13 & \textit{Mangifera indica}         & 85.42 & 93.18 & 89.13 & 88 \\
12 & \textit{Litchi chinensis}         & 98.73 & 75.00 & 85.25 & 104 \\
\midrule
\multicolumn{2}{@{}l}{\textbf{Macro average}} & \textbf{96.11} &
\textbf{96.22} & \textbf{96.04} & 2{,}149 \\
\multicolumn{2}{@{}l}{\textbf{Weighted average}} & \textbf{96.10} &
\textbf{95.91} & \textbf{95.88} & 2{,}149 \\
\bottomrule
\end{tabular}
\end{table}

\subsection{Error structure}
\label{subsec:res-errors}

\begin{figure}[!t]
  \centering
  \includegraphics[width=\columnwidth]{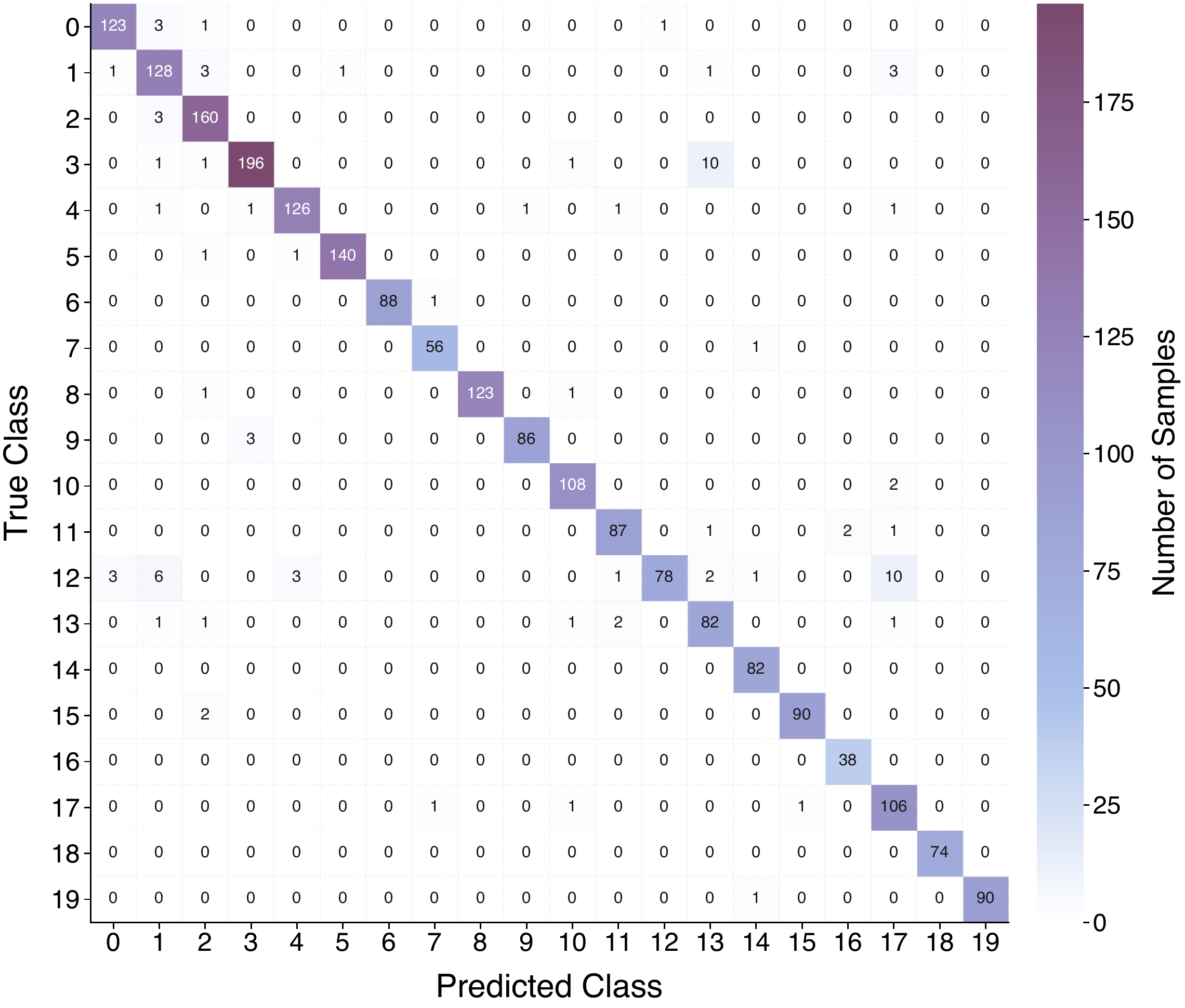}
  \caption{Confusion matrix of the \nameModel reference run.}
  \label{fig:cm}
\end{figure}

The 88 errors are far from uniform (Figure~\ref{fig:cm}): twelve species pairs account for most of the off-diagonal mass, and every dominant pair is botanically interpretable. The largest is \textit{Litchi chinensis} $\rightarrow$ \textit{Monoon longifolium} (9.6\% of lychee images, $n = 10$), both developing smooth dark grey-brown bark that at 20--60\,cm can lack diagnostic structure; remaining lychee errors disperse across \textit{Delonix regia} (5.8\%), \textit{Cocos nucifera} and \textit{Artocarpus heterophyllus} (2.9\% each), so lychee leaks in many directions without attracting anything. \textit{Swietenia macrophylla} $\rightarrow$ \textit{Mangifera indica} (4.8\%, $n = 10$) and \textit{Melaleuca viminalis} $\rightarrow$ \textit{Swietenia macrophylla} (3.4\%) extend a dark-barked dicot cluster, while a smooth-barked cluster links \textit{Delonix regia} to \textit{Khaya anthotheca} and \textit{Monoon longifolium} (2.2\% each). The errors thus concentrate among species genuinely hard to separate from a single patch, in small stable clusters that targeted data (mature lychee and mahogany trunks, wider-standoff views) would plausibly address without architectural change.

\subsection{Comparison with modern backbones}
\label{subsec:res-baselines}

\begin{figure*}[!t]
  \centering
  \includegraphics[width=0.85\textwidth]{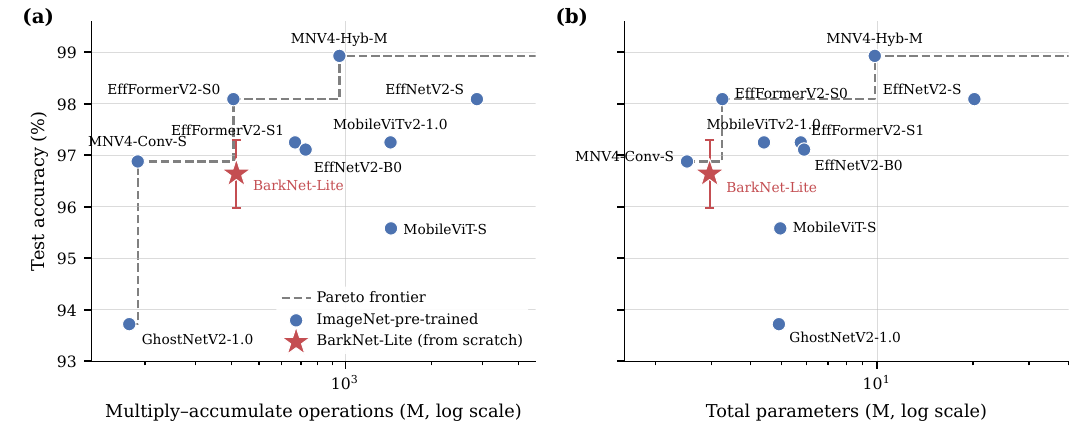}
  \caption{Accuracy against (a) compute and (b) parameters on \nameDataset.}
  \label{fig:frontier}
\end{figure*}

Table~\ref{tab:baselines} compares \nameModel against nine ImageNet-pre-trained backbones fine-tuned on \nameDataset. As an accuracy ranking it places eighth of ten, its five-seed mean 2.3 points below the strongest baseline (MobileNetV4-Hybrid-Medium, 98.93\% at 9.82\,M). As an accuracy--cost frontier the picture improves: it is the second smallest entry (2.96\,M, 11.32\,MB), the only one trained from scratch, and stays ahead of GhostNetV2-100 (93.72\%, 4.90\,M) and MobileViT-S (95.58\%, 4.95\,M), both larger \emph{and} pre-trained. The most direct comparison, MobileNetV4-Conv-Small (2.52\,M, the only other sub-3\,M model), reaches 96.88\%, inside one seed-level standard deviation of our $96.64 \pm 0.66\%$, with ImageNet initialisation we do not require, while the next up, EfficientFormerV2-S0 (3.25\,M), reaches 98.09\%. Compute is less flattering: at 416.52\,M MACs \nameModel is only fourth cheapest, behind GhostNetV2-100 (176.32\,M), MobileNetV4-Conv-Small (188.73\,M) and EfficientFormerV2-S0 (406.69\,M). Parameters and compute decouple---it executes 140.7 MACs per parameter, near the 142.2 of EfficientNetV2-S and nearly four times the 35.97 of GhostNetV2-100---because the \trb bottleneck holds parameters down while the six \tim branches run at full stage resolution. It therefore does not lie on the accuracy--compute Pareto frontier, where MobileNetV4-Conv-Small dominates it outright (0.85$\times$ parameters, 0.45$\times$ compute, 0.85$\times$ disk, 0.24 points higher; Figure~\ref{fig:frontier}). Two caveats bound every comparison in this table. Each baseline is a single fine-tuning run, so none carries an interval and gaps smaller than roughly 1.3 points---the resolution set by our own seed spread---should not be read as ordering. And because the baselines are pre-trained while \nameModel is not, the table measures the joint cost of compact design and ImageNet independence; it cannot attribute the gap to either alone.

\begin{table*}
\centering
\caption{\nameModel against nine ImageNet-pre-trained backbones on \nameDataset, each a single fine-tuning run under the identical protocol. The two \nameModel rows give the reference run and the five-seed mean; the final row is the min--max baseline-to-\nameModel ratio per axis.}
\label{tab:baselines}
\scriptsize
\setlength{\tabcolsep}{2.5pt}
\renewcommand{\arraystretch}{1.12}
\begin{tabular}{@{}l r r r r@{}}
\toprule
\textbf{Model} & \textbf{Params} & \textbf{MACs} & \textbf{Size} & \textbf{Acc.} \\
 & \textbf{(M)} & \textbf{(M)} & \textbf{(MB)} & \textbf{(\%)} \\
\midrule
MobileNetV4-Hybrid-M \cite{qin2024mobilenetv4} & 9.82 & 952.70 & 37.46 & 98.93 \\
EfficientNetV2-S \cite{tan2021efficientnetv2}  & 20.20 & 2873.02 & 77.07 & 98.09 \\
EfficientFormerV2-S0 \cite{li2023rethinking}   & 3.25 & 406.69 & 12.41 & 98.09 \\
EfficientFormerV2-S1 \cite{li2023rethinking}   & 5.74 & 667.39 & 21.91 & 97.25 \\
MobileViTv2-100 \cite{mehta2023separable}      & 4.40 & 1436.27 & 16.78 & 97.25 \\
EfficientNetV2-B0 \cite{tan2021efficientnetv2} & 5.88 & 726.41 & 22.45 & 97.11 \\
MobileNetV4-Conv-S \cite{qin2024mobilenetv4}   & 2.52 & 188.73 & 9.61 & 96.88 \\
MobileViT-S \cite{mehta2022mobilevit}          & 4.95 & 1441.33 & 18.88 & 95.58 \\
GhostNetV2-100 \cite{tang2022ghostnetv2}       & 4.90 & 176.32 & 18.70 & 93.72 \\
\midrule
\textbf{\nameModel (ours)}   & \textbf{2.96} & \textbf{416.52} & \textbf{11.32} &
\textbf{95.91} \\
\textbf{\nameModel, 5 seeds} & \textbf{2.96} & \textbf{416.52} & \textbf{11.32} &
$\mathbf{96.64 \pm 0.66}$ \\
\textit{ratio (min--max)} & \textit{0.85--6.82$\times$} &
\textit{0.42--6.90$\times$} & \textit{0.85--6.81$\times$} & --- \\
\bottomrule
\end{tabular}
\end{table*}

\subsection{Deployment efficiency}
\label{subsec:res-efficiency}

What decides field usability is on-device behaviour, which parameters and MACs predict only loosely \cite{howard2019searching,krishnamoorthi2018quantizing}. \emph{Conversion is lossless; quantisation is not.} The single-precision TensorFlow Lite graph reproduces the Keras test accuracy of 95.91\% exactly, but dynamic-range and full-integer quantisation fall to 84.46\% and 84.27\%---losses of 11.45 and 11.63 points (Table~\ref{tab:deploy})---so a 3.7-fold file-size reduction is not worth eleven points on a twenty-way problem and we do not report a quantised deployment. The two quantised variants differ by only 0.19 points and dynamic-range quantisation leaves activations in floating point, so the damage lies in the weights: the architecture contains the operators most implicated in weight-quantisation failure, the depthwise \cam whose per-channel weight ranges differ by orders of magnitude \cite{krishnamoorthi2018quantizing,nagel2019data} and the six-branch \tim concatenated before one projection. This is a diagnosis rather than a demonstration; the standard remedies \cite{nagel2019data,jacob2018quantization} were not applied, so these figures bound what compression is achievable rather than measure it.

\begin{figure*}[!t]
  \centering
  \includegraphics[width=0.9\textwidth]{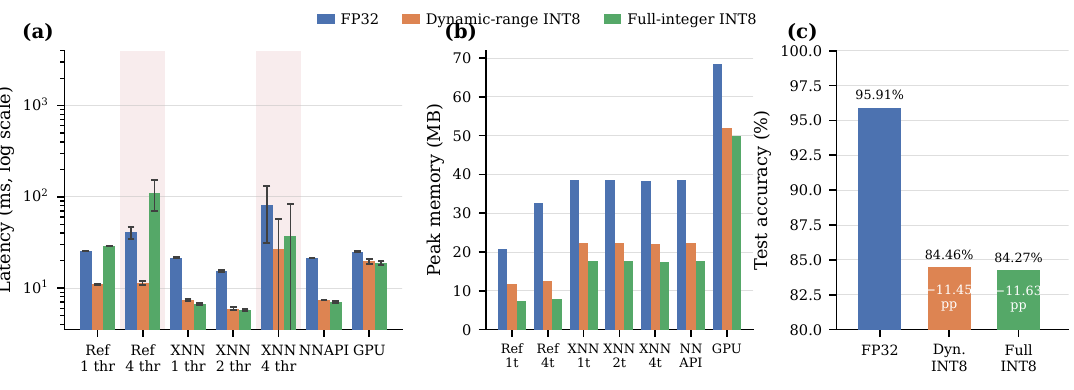}
  \caption{On-device (a) latency, (b) peak memory and (c) accuracy of the three exported artefacts.}
  \label{fig:ondevice}
\end{figure*}

\emph{The delegate matters more than the arithmetic.} Across seven configurations (Figure~\ref{fig:ondevice}) the full-integer graph is the \emph{slowest} under reference kernels (29.09\,ms on one thread against 25.43\,ms for single precision) yet drops to 6.70\,ms under XNNPACK, a factor of 4.34; NNAPI matches XNNPACK to within 0.31\,ms, indicating no accelerator was engaged. Four-thread runs are bimodal (standard deviations 61--121\% of the mean against $\le 3.6\%$ at one and two threads), the signature of a thread pool spread across cores of unequal frequency (1901, 2611, 2803\,MHz), so we report two threads. The GPU delegate is slower on every variant (18.71--24.89\,ms), triples peak memory to 49.86--68.46\,MB and costs 0.75--1.04\,s to initialise against 10.0--22.7\,ms.

\emph{The deployable configuration.} Under XNNPACK on two performance cores the single-precision graph classifies one uncropped $224 \times 224$ photograph in $15.34 \pm 0.39$\,ms (65.20 images/s) within 38.58\,MB peak memory after 20.71\,ms initialisation, at the full 95.91\% accuracy (Table~\ref{tab:deploy}); the same graph needs $24.34 \pm 1.11$\,ms on a desktop CPU, so the phone is not the binding constraint. \nameModel is thus deployable as trained, and the compression that would have made it dramatically cheaper (2.68-fold to 5.71\,ms, 2.19-fold memory to 17.62\,MB) is exactly what post-training quantisation fails to deliver.

\begin{table}[!t]
\centering
\caption{Exported \nameModel artefacts and their measured behaviour on a Snapdragon 7+ Gen 3 smartphone. $\Delta$ is the accuracy change against the single-precision graph.}
\label{tab:deploy}
\scriptsize
\setlength{\tabcolsep}{3pt}
\renewcommand{\arraystretch}{1.12}
\begin{tabular}{@{}l r r r r r@{}}
\toprule
\textbf{Artefact} & \textbf{Size} & \textbf{Acc.} & $\boldsymbol{\Delta}$ &
\textbf{Latency} & \textbf{Peak} \\
 & \textbf{(MB)} & \textbf{(\%)} & \textbf{(pp)} & \textbf{(ms)} & \textbf{(MB)} \\
\midrule
Keras, FP32\textsuperscript{a} & 11.32 & 95.91 & --- & --- & --- \\
\midrule
\textbf{TFLite, FP32} & \textbf{11.32} & \textbf{95.91} & \textbf{0.00} &
$\mathbf{15.34 \pm 0.39}$ & \textbf{38.58} \\
TFLite, dynamic-range INT8 & 2.98 & 84.46 & $-11.45$ & $5.91 \pm 0.21$ & 22.26 \\
TFLite, full-integer INT8 & 3.04 & 84.27 & $-11.63$ & $5.71 \pm 0.14$ & 17.62 \\
\bottomrule
\end{tabular}

\vspace{3pt}
\raggedright\footnotesize
\textsuperscript{a}\,Training artefact, evaluated on the host rather than on device;
listed to show that conversion to the single-precision TFLite graph is lossless.
Throughput for the three TFLite rows is 65.20, 169.19 and 175.00 images per second,
and initialisation 20.71, 10.99 and 10.02\,ms. The deployed configuration is bold.
\end{table}

\subsection{Cross-dataset generalisation}
\label{subsec:res-crossdata}

Retraining the architecture from scratch on three public benchmarks, changing only the classifier width (Table~\ref{tab:crossdata}), transfer is strong where the target resembles \nameDataset in scale and acquisition style and weak where it does not. On BarkVN-50 it reaches 95.86\% across 50 species, matching its own \nameDataset performance and exceeding the 94\% reported for a fine-tuned ResNet-101 \cite{faizal2022automated} with roughly one-fifteenth the parameters and no pre-training; BarkVN-50 ships no official partition, so that figure was obtained on our own stratified split and the comparison is indicative rather than like-for-like. On BarkNet 1.0 it reaches 92.85\% across 23 species, comparable to the 93.88\% single-image figure of \cite{carpentier2018tree}, which reaches 97.81\% only after majority vote over all images of a tree, an option our single-image protocol forgoes. Performance falls sharply on TRUNK12 (69.76\%), which offers only 393 images across 12 classes (about 33 per species against 713 in \nameDataset)---too few for a from-scratch network to build a texture representation, the regime where pre-training or a handcrafted descriptor is the right tool, and a boundary condition on the design rather than a defect.

\begin{table}[!t]
\centering
\caption{Generalisation of the \nameModel architecture, retrained from scratch on each corpus. The last column reports the figure given by the cited work; it is not necessarily the strongest published result on that corpus, and the partition is identical to ours only where the corpus ships one.}
\label{tab:crossdata}
\footnotesize
\setlength{\tabcolsep}{3.5pt}
\renewcommand{\arraystretch}{1.12}
\begin{tabular}{@{}l c r r r l@{}}
\toprule
\textbf{Dataset} & \textbf{Cls.} & \textbf{Acc.} & \textbf{Prec.} &
\textbf{F1} & \textbf{Published} \\
 & & \textbf{(\%)} & \textbf{(\%)} & \textbf{(\%)} & \textbf{ref.\ (\%)} \\
\midrule
\nameDataset (ours) & 20 & 95.91 & 96.10 & 95.88 & -- \\
BarkVN-50 \cite{faizal2022automated} & 50 & 95.86 & 96.44 & 96.38 &
94.0\textsuperscript{a} \\
BarkNet 1.0 \cite{carpentier2018tree} & 23 & 92.85 & 93.04 & 92.86 &
93.9\,/\,97.8\textsuperscript{b} \\
TRUNK12 \cite{svab2014trunk12} & 12 & 69.76 & 74.65 & 66.33 & -- \\
\bottomrule
\end{tabular}

\vspace{3pt}
\raggedright\footnotesize
\textsuperscript{a}\,Single image, ImageNet-pre-trained backbone.
\textsuperscript{b}\,Single image / majority vote over all images of a tree.
\end{table}

\subsection{Ablation study}
\label{subsec:res-ablation}

\begin{figure*}[!t]
  \centering
  \includegraphics[width=0.82\textwidth]{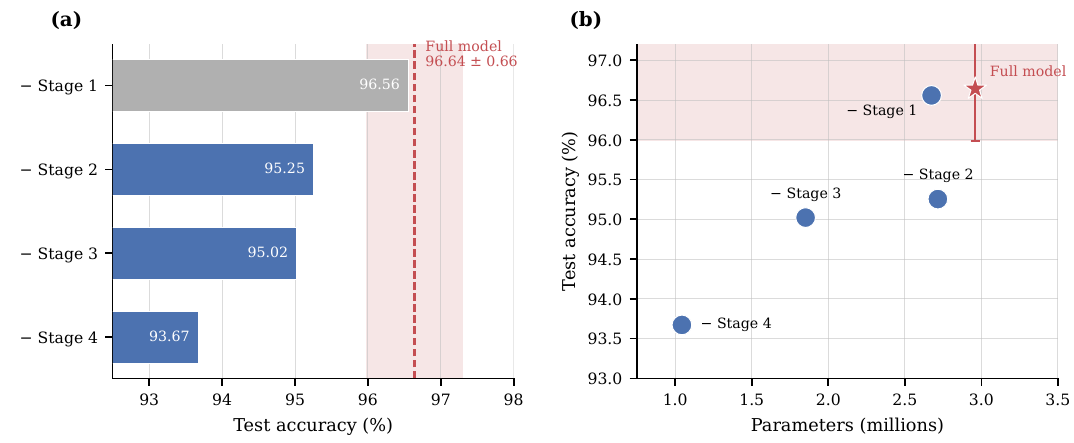}
  \caption{Stage-level ablation across the four single-stage removals.}
  \label{fig:ablation_stage}
\end{figure*}

Removing one whole texture stage at a time and retraining from scratch, read against parameter count (Table~\ref{tab:ablation}, Figure~\ref{fig:ablation_stage}), the ordering is monotonic in depth. Every variant runs for a fixed 100 epochs while the full model trains to a validation plateau at a mean best epoch of 216 (Table~\ref{tab:multiseed}), so the variants are budget-matched to one another but not to it; each reported drop is therefore an upper bound on the cost of that removal, and the ordering rather than the point value is what the analysis rests on. Removing the deepest stage is most damaging (93.67\% at 1.04\,M parameters, 2.97 points below the five-seed model), the third costs 1.62 points, the second 1.39, and the first essentially nothing; McNemar's test \cite{mcnemar1947note} confirms the three deeper removals are each significantly worse than the mildest (Stage~4 $p = 3 \times 10^{-10}$, Stage~3 $p = 2 \times 10^{-4}$, Stage~2 $p = 4 \times 10^{-4}$). Capacity does not explain the ordering---removing Stage~1 and Stage~2 leave almost identical parameter counts (2.67 and 2.72\,M) yet cost 0.08 and 1.39 points---so which stage remains matters more than how many parameters remain, and the deepest stage carries the most class-discriminative signal, consistent with the Grad-CAM evidence of Section~\ref{subsec:res-xai}. The first-stage removal (96.56\%, 0.08 below the mean at 10\% fewer parameters) is within seed noise and is confounded with the colour pathway that taps the same stage, so it can be attributed to neither alone; we keep the full architecture as reference and rely on the export path of Section~\ref{subsec:res-efficiency} rather than architectural trimming. Per class, removing the deepest stage costs most on the three finest-grained surfaces---\textit{Mimusops elengi} (class~8), \textit{Areca catechu} (class~5) and \textit{Delonix regia} (class~1), which lose between 9 and 17 F1 points---consistent with that stage supplying the fine texture they depend on.

\begin{table}[!t]
\centering
\caption{Stage-level ablation of \nameModel across the four single-stage removals.}
\label{tab:ablation}
\footnotesize
\setlength{\tabcolsep}{3.5pt}
\renewcommand{\arraystretch}{1.15}
\begin{tabular}{@{}l r r r r l@{}}
\toprule
\textbf{Variant} & \textbf{Params} & \textbf{Acc.} & \textbf{F1} &
$\boldsymbol{\Delta}$ & \textbf{McNemar} \\
 & \textbf{(M)} & \textbf{(\%)} & \textbf{(\%)} & \textbf{(pp)} & \textbf{$p$} \\
\midrule
Full model (5-seed) & 2.96 & $96.64$ & $96.61$ & --- & --- \\
\midrule
$-$Stage 1\textsuperscript{a} & 2.67 & $96.56$ & $96.55$ & $-0.08$ & ref. \\
$-$Stage 2 & 2.72 & $95.25$ & $95.27$ & $-1.39$ & $4\times10^{-4}$ \\
$-$Stage 3 & 1.85 & $95.02$ & $95.04$ & $-1.62$ & $2\times10^{-4}$ \\
$-$Stage 4 & 1.04 & $93.67$ & $93.76$ & $-2.97$ & $3\times10^{-10}$ \\
\bottomrule
\end{tabular}

\vspace{2pt}
\raggedright\footnotesize
\textsuperscript{a}\,The recorded configuration removes the first texture stage
together with the colour pathway that taps from it, so this row confounds the two;
its near-zero effect cannot be attributed to either alone. Full-model row is the
five-seed mean of Section~\ref{subsec:res-multiseed}; ablation rows are single
seeds at 100 epochs.
\end{table}

\subsection{Explainability}
\label{subsec:res-xai}

An accurate classifier trained on uncropped, site-clustered images may be right for a reason that will not survive deployment, so we ground RQ4 in a fixed qualitative panel plus three quantitative probes: deletion and insertion faithfulness \cite{petsiuk2018rise}, Grad-CAM++ average-drop and increase-in-confidence \cite{chattopadhay2018grad}, and the weight-randomisation sanity check \cite{adebayo2018sanity}.

\begin{figure*}[!t]
  \centering
  \includegraphics[width=\textwidth]{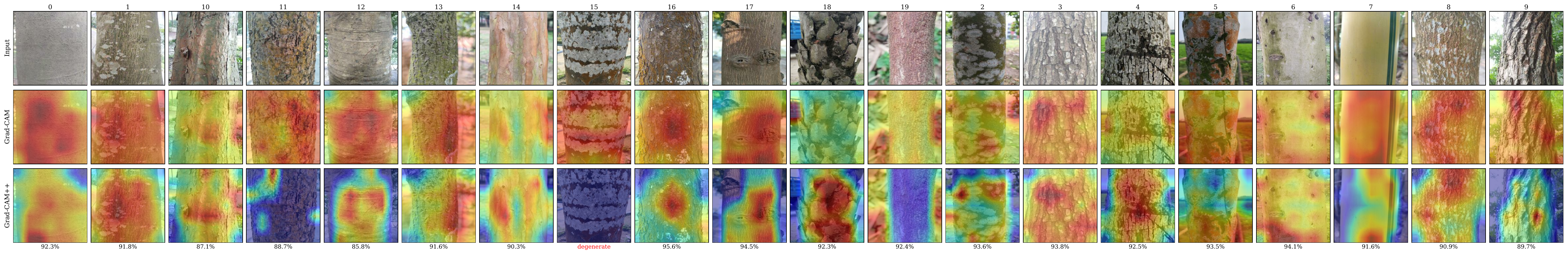}
  \caption{Grad-CAM (middle) and Grad-CAM++ (bottom) maps of the reference run, for one correct test image per species.}
  \label{fig:gradcam_grid}
\end{figure*}

\emph{(i) Maps fall on bark, not background.} Across all twenty classes (Figure~\ref{fig:gradcam_grid}) the activation mass concentrates on the trunk rather than foliage, sky or litter, so the model has not learned the site shortcut that was the principal risk of collecting at few locations \cite{beery2018recognition}, and within the trunk it lands on nameable features: fissure lines, lenticel fields, leaf-base scar bands. This matches Kim et al.~\cite{kim2022identifying} on 42 temperate species; one class (\textit{Roystonea regia}) gave a degenerate near-uniform map. \emph{(ii)} The attribution refines with depth, figure--ground separation emerging only in the deepest two stages (Supplementary Figure~S2), consistent with the ablation. \emph{(iii) The failures are of discrimination, not attention:} for misclassified images the true- and predicted-class maps fall on the same bark patch (Supplementary Figure~S3); over all 88 errors the mean probability is 0.55 on the wrong class and 0.20 on the true class, and 22 still place above 0.30 on the true class; the true species is the second choice often enough that top-2 accuracy reaches 98.88\% (Section~\ref{subsec:res-calibration}), so a two-candidate shortlist recovers roughly three quarters of these errors at no cost, and the remedy is finer resolution, not stronger attention. \emph{(iv) Faithfulness.} Over 200 images (Table~\ref{tab:faithfulness}, Supplementary Figure~S4) Grad-CAM separates clearly from a random control (deletion AUC $0.298 \pm 0.154$ against 0.342, insertion $0.476 \pm 0.117$ against 0.354); the separation is modest, the expected signature of a texture classifier whose evidence spreads across the trunk, and Grad-CAM is more faithful than Grad-CAM++ on every measure despite the latter's sharper maps, so all qualitative panels use Grad-CAM. \emph{(v) Sanity check.} Cascade-randomising the classifier head and final stage \cite{adebayo2018sanity} collapses the maps to a near-uniform field (Supplementary Figure~S5) that does not revert to stage~I edge structure, so the localisation depends on the learned weights. The $7 \times 7$ grid resolves the discriminative region but not individual fissures, and the AUCs are meaningful against the control rather than absolutely, so the RQ4 claim rests on the deletion, insertion and randomisation evidence together.

\begin{table}[!t]
\centering
\caption{Faithfulness of the attribution maps over 200 \nameDataset test images.}
\label{tab:faithfulness}
\footnotesize
\setlength{\tabcolsep}{3pt}
\renewcommand{\arraystretch}{1.15}
\begin{tabular}{@{}l c c r r@{}}
\toprule
\textbf{Method} & \textbf{Del.\ AUC} & \textbf{Ins.\ AUC} &
\textbf{Avg.\ drop} & \textbf{Increase} \\
 & $\boldsymbol{\downarrow}$ & $\boldsymbol{\uparrow}$ &
\textbf{(\%) }$\boldsymbol{\downarrow}$ & \textbf{(\%) }$\boldsymbol{\uparrow}$ \\
\midrule
Random (control) & 0.342 & 0.354 & --- & --- \\
Grad-CAM++ \cite{chattopadhay2018grad} & 0.306 & 0.454 & 35.21 & 10.00 \\
\textbf{Grad-CAM} \cite{selvaraju2017grad} & \textbf{0.298} & \textbf{0.476} &
\textbf{6.64} & \textbf{36.00} \\
\bottomrule
\end{tabular}
\end{table}

\subsection{Calibration}
\label{subsec:res-calibration}

Correct and incorrect predictions separate clearly by confidence (Supplementary Figure~S6), so confidence is usable as a rejection signal, and top-$k$ accuracy rises from 95.91\% at $k = 1$ to 98.88\% at $k = 2$ and 99.44\% at $k = 3$. The reliability diagram gives an expected calibration error of 0.148 over 15 equal-width bins and shows systematic under-confidence, the expected signature of label smoothing, which with $\varepsilon = 0.05$ and $C = 20$ caps the target probability at 0.9525 \cite{muller2019does,guo2017calibration}; the direction is benign, and a single temperature-scaling parameter fitted on validation would remove most of the gap.


\section{Discussion}
\label{sec:discussion}

\subsection{Answering the research questions}
\label{subsec:disc-rqs}

\textbf{RQ1: the accuracy--cost frontier.} At $96.64 \pm 0.66\%$ with 2.96\,M parameters and no pre-training, \nameModel trails the strongest fine-tuned backbones by up to 2.3 points yet sits within one seed-level standard deviation of MobileNetV4-Conv-Small, the only smaller entry in the table. The picture at that end of the size range is not uniform: EfficientFormerV2-S0, larger by 10\%, reaches 98.09\%, so a from-scratch design matches some compact pre-trained backbones at this budget and not others. It is not on the accuracy--compute frontier, however, spending compute where it saves parameters (140.7 MACs per parameter). What survives is a measured claim---one photograph in $15.34 \pm 0.39$\,ms within 38.58\,MB on a mid-range phone at full accuracy---and the gap is the price of ImageNet independence, worth paying when checkpoint licensing is restrictive, the domain is far from ImageNet, or provenance must be auditable.

\textbf{RQ2: transfer.} Retrained from scratch the architecture reaches 95.86\% on BarkVN-50 and 92.85\% on BarkNet 1.0, close to the single-image results reported for much larger fine-tuned backbones \cite{faizal2022automated,carpentier2018tree}, so performance is not specific to our species or conditions. TRUNK12 (69.76\%) delimits the claim: 33 images per class cannot support a texture representation learned from scratch, the regime transfer learning was made for, which our design concedes.

\textbf{RQ3: which components carry performance.} The ablation is monotonic in depth---the deepest stage costs 2.97 points to remove, the first essentially nothing---and is not explained by capacity, since near-equal parameter counts give very different losses, so the deepest stage carries the most class-discriminative signal. Being single-seed and trained on a shorter budget than the full model, it cannot resolve the near-zero first-stage effect against the 0.66-point noise floor, and that removal is confounded with the colour pathway tapping the same stage; separating the two remains future work.

\textbf{RQ4: grounding in bark.} Grad-CAM mass falls on trunk surfaces and concentrates on fissures, lenticel fields and leaf-base scars, matching expert diagnostic keys and the pattern of \cite{kim2022identifying}; deletion and insertion curves separate from a random control, and weight randomisation confirms the localisation is a property of the learned parameters. The failures are of discrimination rather than attention: the model localises the correct region and then assigns it to the wrong species with high confidence, so the remedy is finer input resolution.

\subsection{Implications}
\label{subsec:disc-implications}

Five points extend beyond this dataset. First, \emph{regional datasets are the binding constraint, not architectures}: no algorithmic advance could recognise Bangladeshi bark while the species were absent from every label space, so our acquisition protocol, not our network, is the transferable contribution. Second, \emph{compactness can be designed rather than retrofitted, at a price}: where Deep BarkID \cite{wu2021deep} distilled a large pre-trained model, designing around the signal reaches a comparable operating point at 2.96\,M parameters with no external dependency, for 2.3 points. Third, \emph{parameter count is not a cost metric}: \nameModel is second smallest on parameters yet only fourth on compute, so MACs and measured latency must both be reported. Fourth, \emph{verification belongs in the reporting}: an uncropped, site-clustered dataset risks learning location rather than species, and the Grad-CAM analysis is what licenses the deployment claim. Finally, the uncorrelated per-species F1 and image count ($r = -0.12$) shifts the effort from volume to acquisition protocol, and the strong top-$k$ behaviour argues for a shortlist interface over a single verdict.

\subsection{Limitations and threats to validity}
\label{subsec:disc-limitations}

\emph{Evaluation split.} This is the largest threat to external validity. The released partition is stratified by image, and since each tree contributes 10 to 40 frames, photographs of one trunk fall on both sides of it; our accuracies therefore measure recognition of the sampled individuals under unseen viewpoints, distances and illumination rather than generalisation to unseen individuals, for which a tree-grouped partition is the right test \cite{carpentier2018tree} and under which every figure here is an upper bound, plausibly a loose one. The constraint applies equally to the nine baselines, which share the split, so the \emph{relative} comparison is unaffected. \emph{Baselines and ablation.} Each backbone was fine-tuned once, so only our own side of the comparison carries an interval; and because the baselines are pre-trained while \nameModel is not, the 2.3-point gap measures compact design and ImageNet independence jointly rather than architecture alone. The stage removals ran for 100 epochs against a full model trained to a plateau at a mean best epoch of 216, so their deltas bound each removal from above and only the ordering is claimed. \emph{Quantisation.} Both quantised exports lose more than eleven points, leaving the 11.32\,MB single-precision graph as the deployable artefact; we localised the loss to the weights but not to the responsible layer. \emph{Single runs and single device.} Cross-dataset numbers are one run per corpus, and on-device latency and memory were measured on one smartphone, so figures on other silicon will differ; energy and thermal behaviour were not measured. \emph{Scope.} Collection covered four districts over a limited period and sought diameter variation without recording it, so generalisation across regions, seasons and tree ages---the variable \cite{warner2024centralbark} found only weakly correlated with accuracy---is untested. \emph{Attribution and calibration.} The $7 \times 7$ grid cannot separate the confident smooth-barked confusions, the faithfulness AUCs are meaningful only against the control, and the model is benignly under-confident, so any deployment surfacing confidence should apply temperature scaling fitted on validation.

\subsection{Future work}
\label{subsec:disc-future}

Two priorities follow directly from the limitations: a tree-grouped re-partition of \nameDataset, which would convert the upper bound above into a measurement, and a multi-seed ablation separating the colour pathway from the first texture stage. Beyond those: recorded trunk diameters, to measure ontogenetic robustness; a multi-distance protocol capturing a close texture patch and a wider trunk view together, turning the smooth-barked confusions into a data fix; per-tree aggregation over the several images a user naturally takes, which lifted BarkNet results by almost four points \cite{carpentier2018tree}; and recovering the quantisation loss through quantisation-aware training or per-channel range equalisation \cite{nagel2019data,jacob2018quantization}, bringing a 3.04\,MB, 5.71\,ms artefact within reach.


\section{Conclusion}
\label{sec:conclusion}

Bark is the one diagnostic organ available to a field surveyor in every season and at every tree height, yet bark recognition has been developed almost entirely for temperate floras and on large ImageNet-pre-trained backbones. This paper addressed both limitations. We introduced \nameDataset, the first bark image dataset for Bangladesh: 14{,}258 uncropped smartphone photographs of 20 native hardwood and palm species across four districts under sunny, cloudy and rainy conditions, with English, Bengali and scientific names per class and a fixed stratified split. On it we proposed \nameModel, a 2.96\,M-parameter network trained from random initialisation that pairs a multi-scale texture pathway with a parallel colour-aware pathway, reaching $96.64 \pm 0.66\%$ accuracy over five seeds under strict single-image inference.

Five findings carry beyond the specific model. The accuracy--cost trade-off is real but modest: against nine compact ImageNet backbones \nameModel sits within one seed-level standard deviation of the smallest and 2.29 points below the best, though on neither Pareto frontier, so independence from pre-training is affordable rather than free. It is nonetheless deployable as trained, classifying one photograph in $15.34 \pm 0.39$\,ms within 38.58\,MB on a mid-range smartphone, while post-training quantisation costs over eleven points and is unusable without quantisation-aware training. Transfer is bounded by data regime rather than flora: 95.86\% on BarkVN-50 and 92.85\% on BarkNet 1.0, but 69.76\% on the 33-image-per-class TRUNK12. Per-species accuracy is uncorrelated with image count ($r = -0.12$), so what limits performance is intrinsic bark separability, redirecting future collection from volume towards acquisition protocol. Finally, the residual errors form botanically coherent clusters, and deletion, insertion and weight-randomisation checks confirm that predictions rest on fissures, lenticel fields and leaf-base scars rather than background.

These figures come from an image-level split, so they bound recognition of the sampled trees rather than of unseen individuals; the qualification does not change the comparisons, which share the split, but it sets what a field deployment should expect.

Above all, the geographic concentration of bark research is a data problem before it is an algorithmic one: no architecture could have produced a Bangladeshi bark recogniser while the species were absent from every available label space. Building regionally grounded datasets, and pairing them with models small enough to run on the devices surveyors already carry, is what determines whether automated species recognition reaches the regions that need it most.


\section*{Data and code availability}

The \nameDataset dataset, the training and evaluation code, the trained \nameModel weights and the figure-generation notebooks will be released at \url{https://github.com/aroshi2984/BarkBD} upon acceptance, and archived with a persistent identifier in a public research-data repository. The release includes the fixed stratified train/validation/test split of Section~\ref{subsec:data-pipeline}, so that subsequent work can report directly comparable figures without re-splitting; the checkpoints of the five seeds of Table~\ref{tab:multiseed} and of the reference run of Section~\ref{subsec:res-overall}; and the signed expert botanical-validation certificate for the \nameDataset labels (Section~\ref{subsec:data-annotation}).


\section*{Declaration of competing interest}

The authors declare that they have no known competing financial interests or personal relationships that could have appeared to influence the work reported in this paper.


\section*{Funding}

This research did not receive any specific grant from funding agencies in the public, commercial, or not-for-profit sectors.


\section*{CRediT authorship contribution statement}

\textbf{Aroshi Ali:} Conceptualization, Methodology, Software, Investigation, Data curation, Writing -- review \& editing.
\textbf{Saad Ahmed:} Conceptualization, Supervision, Project administration, Writing -- review \& editing.
\textbf{Md.\ Khalid Syfullah:} Formal analysis, Validation, Visualization, Writing -- original draft, review \& editing.


\section*{Declaration of generative AI and AI-assisted technologies in the writing process}

During the preparation of this work the authors used Large Language Models (LLM) in order to improve the language and readability of the manuscript. After using this tool, the authors reviewed and edited the content as needed and take full responsibility for the content of the published article. No generative-AI tool was used to produce, analyse or interpret the data, to generate or alter any figure, or to derive any of the numerical results reported here.


\section*{Acknowledgements}

The authors thank Prof.\ Kazi Kamrul Islam (Department of Agroforestry, Faculty of Agriculture, Bangladesh Agricultural University (BAU), Mymensingh-2202, Bangladesh) for verifying the species identity of the sampled trees. The signed validation certificate is retained by the authors and will be released with the \nameDataset dataset in the project repository. We are grateful to the authorities of the public parks, the campus administrations and the homestead owners across Rangpur, Dinajpur, Bogura and Cox's Bazar for granting access during field data collection, and to everyone who assisted with photographing the trees.


\bibliographystyle{elsarticle-num}
\bibliography{references}

\end{document}